\documentclass[sigconf]{acmart}

\copyrightyear{2023} 
\acmYear{2023} 
\setcopyright{acmlicensed}\acmConference[KDD '23]{Proceedings of the 29th ACM SIGKDD Conference on Knowledge Discovery and Data Mining}{August 6--10, 2023}{Long Beach, CA, USA}
\acmBooktitle{Proceedings of the 29th ACM SIGKDD Conference on Knowledge Discovery and Data Mining (KDD '23), August 6--10, 2023, Long Beach, CA, USA}
\acmPrice{15.00}
\acmDOI{10.1145/3580305.3599515}
\acmISBN{979-8-4007-0103-0/23/08}

\AtBeginDocument{%
  }

\usepackage{graphicx}
\usepackage{subcaption}
\usepackage{amsmath}
\usepackage{multirow}
\usepackage{algpseudocode}
\usepackage{algorithm}
\usepackage{tabularx}
\usepackage{enumitem}
\usepackage{pbox}
\usepackage{float}
\usepackage{hyperref}
\usepackage{setspace}
\usepackage{lipsum}

\hypersetup{
    colorlinks=true,
    linkcolor=black,
    filecolor=magenta,      
    urlcolor=cyan,
    pdftitle={Overleaf Example},
    pdfpagemode=FullScreen,
    }
\algrenewcommand\algorithmicrequire{\textbf{Input:}}
\algrenewcommand\algorithmicensure{\textbf{Output:}}

\newcommand{\proposed}{\textsf{TEG}}
\newcommand*{\Comb}[2]{{}^{#1}C_{#2}}%
\begin{document}

\title{Task-Equivariant Graph Few-shot Learning}


\author{Sungwon Kim}
\affiliation{%
  \institution{KAIST}
  \city{Daejeon}
  \country{Republic of Korea}}
\email{swkim@kaist.ac.kr}

\author{Junseok Lee}
\affiliation{%
  \institution{KAIST}
  \city{Daejeon}
  \country{Republic of Korea}}
\email{junseoklee@kaist.ac.kr}

\author{Namkyeong Lee}
\affiliation{%
  \institution{KAIST}
  \city{Daejeon}
  \country{Republic of Korea}}
\email{namkyeong96@kaist.ac.kr}

\author{Wonjoong Kim}
\affiliation{%
  \institution{KAIST}
  \city{Daejeon}
  \country{Republic of Korea}}
\email{wjkim@kaist.ac.kr}

\author{Seungyoon Choi}
\affiliation{%
  \institution{KAIST}
  \city{Daejeon}
  \country{Republic of Korea}}
\email{csyoon08@kaist.ac.kr}

\author{Chanyoung Park}
\authornote{Corresponding author.}
\affiliation{%
  \institution{KAIST}
  \city{Daejeon}
  \country{Republic of Korea}}
\email{cy.park@kaist.ac.kr}
\renewcommand{\shortauthors}{Sungwon Kim et al.} 

\begin{abstract}
Although Graph Neural Networks (GNNs) have been successful in node classification tasks, their performance heavily relies on the availability of a sufficient number of labeled nodes per class.
In real-world situations, not all classes have many labeled nodes and there may be instances where the model needs to classify new classes, making manual labeling difficult.
To solve this problem, it is important for GNNs to be able to classify nodes with a limited number of labeled nodes, known as \textit{few-shot node classification}.
Previous episodic meta-learning based methods have demonstrated success in few-shot node classification, but our findings suggest that optimal performance can only be achieved with a substantial amount of diverse training meta-tasks.
To address this challenge of meta-learning based few-shot learning (FSL), we propose a new approach, the \textbf{T}ask-\textbf{E}quivariant \textbf{G}raph few-shot learning (\proposed) framework.
Our \proposed~framework enables the model to learn transferable task-adaptation strategies using a limited number of training meta-tasks, allowing it to acquire meta-knowledge for a wide range of meta-tasks.
{By incorporating equivariant neural networks, \proposed~ can utilize their strong generalization abilities to learn highly adaptable task-specific strategies.}
As a result, \proposed~ achieves state-of-the-art performance with limited training meta-tasks.
Our experiments on various benchmark datasets demonstrate \proposed's superiority in terms of accuracy and generalization ability, even when using minimal meta-training data, highlighting the effectiveness of our proposed approach in addressing the challenges of meta-learning based few-shot node classification.
Our code is available at the following link: \url{https://github.com/sung-won-kim/TEG}

\end{abstract}

\ccsdesc[500]{Computing methodologies~Artificial intelligence}
\ccsdesc[500]{Computing methodologies~Supervised learning by classification}

\keywords{Node Classification; Few-shot Learning; Graph Neural Networks; Equivariant Neural Networks}


\maketitle

\section{Introduction}
An attributed graph is highly effective at modeling complex networks in the real world, including social networks, biological networks, and transaction networks.
Specifically, node classification is an important task on attributed networks to figure out the characteristics of individual nodes.
For example, it allows us to identify the key influencer on social networks \cite{qi2011exploring, yuan2013latent, kim2022heterogeneous}, find out proteins that play an important role in the biological pathway \cite{zhang2019multimodal,yue2020graph}, or detect fraud by identifying which activities are unusual transactions on the financial transaction network \cite{Li2019ClassifyingAU}.

Although Graph Neural Networks (GNNs) have been successful in node classification tasks \cite{gcn, gat, wu2020comprehensive,gin}, their performance heavily relies on the number of labeled nodes in each class \cite{metagnn,lee2022grafn}. 
However, in practice, not all classes have abundant labeled nodes, and what is even worse is, in many scenarios, we often need to handle nodes that belong to none of the classes that we have seen before, i.e., novel classes, in which case even manually labeling nodes is non-trivial. 
For example, in a biological network, it requires expert domain knowledge to annotate newly discovered protein nodes that are crucial in the biological pathway \cite{zhang2019multimodal,yue2020graph}. 
Hence, it is important for GNNs to be able to perform node classification given a limited amount of labeled nodes, which is known as \textit{few-shot node classification}.

 \begin{figure}[t] 
\begin{center}
\includegraphics[width=1.0\linewidth]{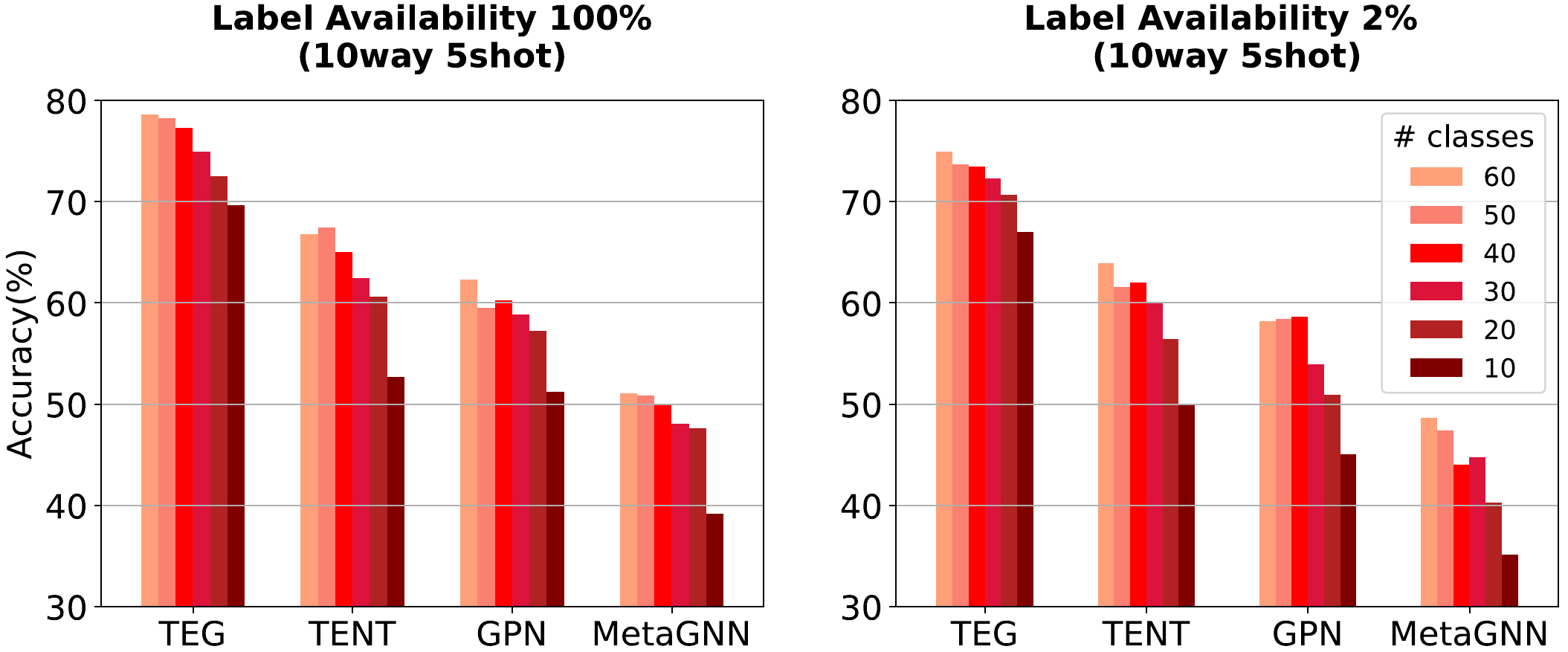}
\end{center}
\caption{Impact of the number of classes used for meta-training when the label availability is 100\% (left) and 2\% (right) on few-shot node classification. Label availability denotes the rate of usable labeled train data for each class. Amazon-electronic dataset is used.}
\label{fig:cls-lbl-test}
\vspace{-5mm}
\end{figure}

{Most studies on the few-shot node classification adopt the episodic meta-learning scheme.} 
Their main idea is to train the model with a series of meta-training tasks each of which contains a small number of labeled nodes (i.e., support nodes) in each class, and a few more labeled nodes in each class for classification (i.e., query nodes). 
This endows the model with the meta-knowledge to classify nodes in novel classes with a few labeled nodes, which is evaluated on meta-test tasks sampled from novel classes. 
As a representative model using an episodic meta-learning paradigm in the graph domain, GPN \cite{gpn} learns a set of prototypes for each class and uses them to classify new instances based on their similarity with the prototypes. 
MetaGNN \cite{metagnn}, a framework that combines attributed network with MAML \cite{maml}, aims to find the initial model parameters that generalize well across all meta-tasks.
G-Meta \cite{gmeta} utilizes subgraphs to generate node representations for performing few-shot node classification. 
Most recent studies focus on addressing the task variance issue caused by different node/class distributions in meta-tasks \cite{gmeta,tadanet}. 
Specifically, these methods involve generating or transforming GNNs' parameters for each meta-task \cite{metagps,tent}, reflecting the relative location between nodes in the meta-task \cite{rale}, or extracting task-specific structures after sampling nodes related to the meta-task \cite{gmeta,glitter}.

Despite the success of existing methods that adopt the episodic meta-learning scheme, we find that they require a large number of training meta-tasks that consist of diverse classes and nodes.
In Figure \ref{fig:cls-lbl-test}, we observe that the performance of TENT \cite{tent}, which is the current state-of-the-art (SOTA) graph few-shot learning (FSL) method, and two representative episodic meta-learning-based-methods, GPN \cite{gpn} and MetaGNN \cite{metagnn}, decreases as the number of classes used for meta-training decreases, i.e., as the diversity of the training meta-tasks decreases.
Moreover, when the label availability is small, i.e., when the diversity of the training meta-tasks is even further decreased, the overall general performance of all methods is greatly reduced.
The above results imply that the \textit{diversity of the training meta-tasks is crucial} for the performance of episodic meta-learning-based methods.
However, in real-world attributed networks, the high cost of labeling often results in limited labeled instances for many classes \cite{gpn}, which makes it difficult to create diverse training meta-tasks that include a sufficient variety of classes and nodes. 
Hence, it is important to develop a method for few-shot node classification that generalizes well given a limited number of training meta-tasks with low diversity.

 \begin{figure}[t] 
\begin{center}
\includegraphics[width=1.0\linewidth]{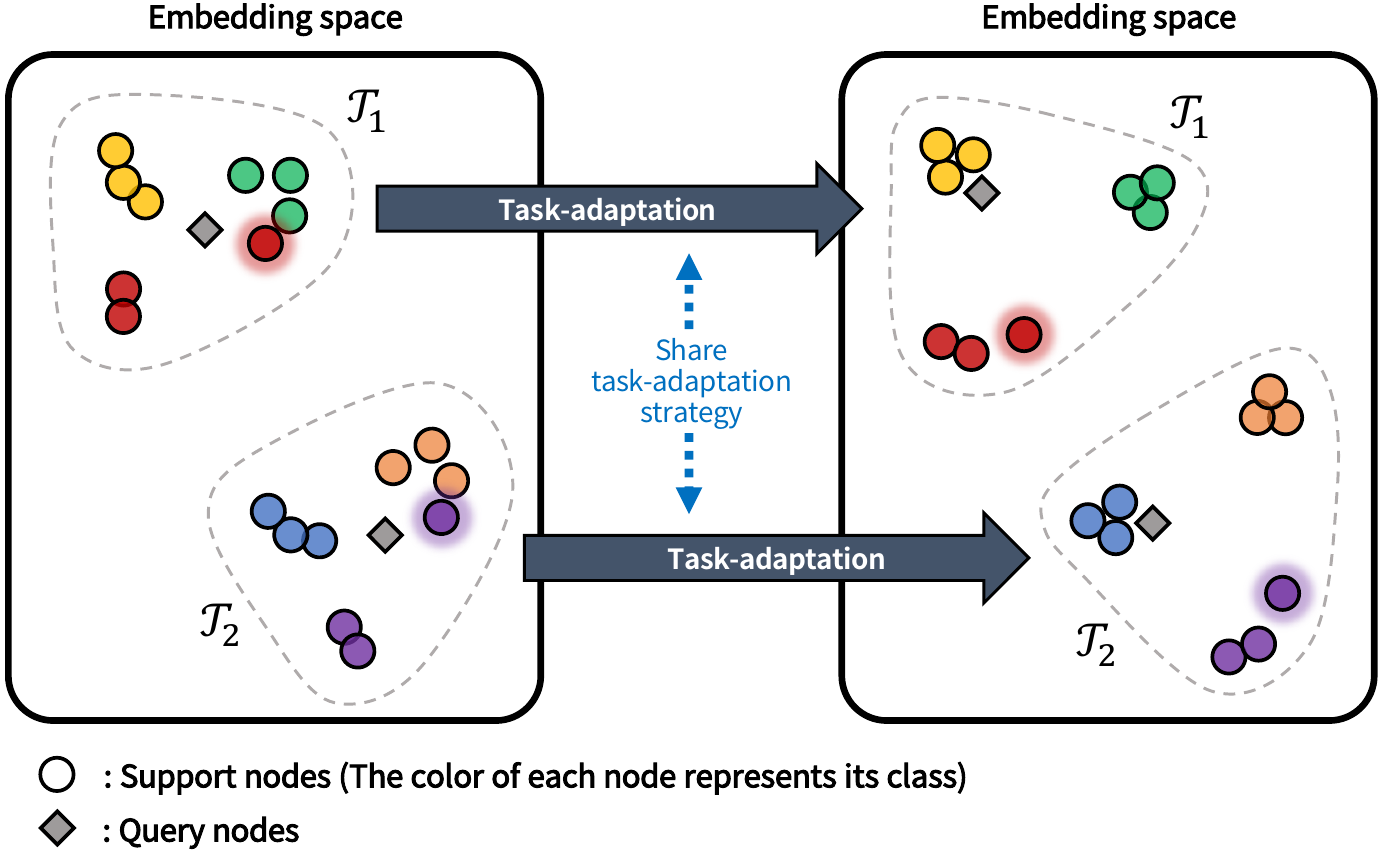}
\end{center}
\caption{
An example of two tasks (i.e., $\mathcal{T}_1$ and $\mathcal{T}_2$) sharing a task-adaptation strategy. A task-adaptation strategy is shared between two tasks whose constituent nodes have similar relational positions between themselves.}
\label{fig:inductivebias}
\vspace{-4.5mm}
\end{figure}

To this end, we propose a novel \underline{\textbf{T}}ask-\underline{\textbf{E}}quivariant \underline{\textbf{G}}raph few-shot learning framework, called \proposed, that learns highly transferable task-adaptation strategies so that the model can acquire meta-knowledge for a wide range of meta-tasks given limited training meta-tasks with low diversity. 
The main idea is to train a task embedder that captures an inductive bias specifying that meta-tasks whose constituent nodes have similar relational positions between themselves in the embedding space (we refer to this as {\textbf{\textit{task-patterns}}}) should share a similar task-adaptation strategy.
To achieve this goal, our task embedder is designed to be {{equivariant to transformations (e.g., rotation, reflection, and translation) of node representations within a meta-task}} (we refer to this as {\textbf{\textit{task-equivariance}}}). 
This facilitates \proposed~to share a similar task-adaptation strategy between two different meta-tasks with similar \textit{task-patterns}, e.g., pulling an outlier node belonging to a particular class (the red shining node in $\mathcal{T}_{1}$ and the purple shining node in $\mathcal{T}_{2}$) closer to the nodes of its own class (See Figure \ref{fig:inductivebias}). 
In other words, \proposed~ is able to share a similar task-adaptation strategy across meta-tasks with similar \textit{task-patterns}, even if the classes and node representations within each meta-task are different, thereby improving generalization well given a limited number of training meta-tasks with low diversity.
On the other hand, as existing models \cite{tent, rale, metagps, glitter, gmeta, metagnn} treat meta-tasks independently, the meta-knowledge should be obtained by learning separate task-adaptation strategies for $\mathcal{T}_{1}$ and $\mathcal{T}_{2}$, and thus require a large number of diverse training meta-tasks for significant performance.

Our main contributions are summarized as follows:  
\begin{enumerate}
    \item We discover that existing episodic meta-learning-based methods require a large number of training meta-tasks that consist of diverse classes and nodes for significant performance.

    \item We propose a novel graph meta-learning framework, called \proposed, for few-shot node classification, that introduces a task embedder that learns highly transferable adaptation strategies. This allows the model to generalize to new meta-tasks with similar \textit{task-patterns}, even if the classes and node representations within each meta-task are different.

    \item Thanks to its strong generalization power, \proposed~outperforms recent state-of-the-art methods, even when the number of train classes and labeled nodes are limited. 
This offsets the disadvantage of the episodic meta-learning scheme, which requires a large and diverse set of training meta-tasks for significant performance.

\end{enumerate}

\section{Preliminaries}
\subsection{Equivariant Neural Networks}
\textit{Equivariance} is a property that defines the output of a network or algorithm corresponding to a symmetric or transformed input.
Let $G$ be some group of transformation (\textit{e.g.} rotation, translation, scaling, permutation etc.) and assume that a mapping function $\phi : \mathcal{X} \rightarrow \mathcal{Y}$ is equivariant to the  $g \in G$ (or $g$-equivariance for short). 
Then, we can represent the input and output of the function as follows:
\begin{equation}
\phi(T_g(x))=T'_g(\phi(x))=T'_g(y)
\end{equation}
where $T_g$ and $T'_g$ are transformations on the input $x\in\mathcal{X}$, and equivalent transformation on its output $y\in\mathcal{Y}$, respectively.
\textit{Invariance} is a special case of equivariance that additionally satisfies $y=T'_g(y)$.

In general, in machine learning, the mapping function $\phi$ is a neural network consisting of non-linear functions.
For example, if the network is equivariant to rotation (i.e., rotation equivariance), the output value of the network can be obtained by rotating as much as the input value is rotated.
Satisfying equivariance to various transformations allows the network to identify patterns or structures contained in the input regardless of its location or direction.
Thanks to these advantages, equivariance is considered an important property in machine learning because it can endow the network with generalizability on new data.
Research on equivariant neural networks has yielded significant advancements in various downstream tasks 
 \cite{cohen2016group, cohen2016steerable, weiler2019general, rezende2019equivariant, romero2020group}, such as N-body systems, 3D point clouds \cite{kohler2019equivariant,kohler2020equivariant} and even the behavior and properties of molecules \cite{anderson2019cormorant, miller2020relevance} with a high level of generalization power.
Moreover, EGNN \cite{egnn} achieves E(n) equivariance (i.e., Euclidean group equivariance in $n$-dimension space) regarding the $n$-dimension coordinate of instances even reflecting {the property of each instance}. 
In contrast to existing EGNN, we use latent features (i.e., node representations) as the coordinates of nodes in the embedding space, instead of their actual coordinates in the real-world Euclidean space. 
This allows us to learn a general metric function for FSL that is based on the relative differences between features of different instances in the same meta-task, rather than the absolute features themselves.

\section{Problem Statement}
\subsubsection*{\textbf{Notations}}  
Let $\mathcal{G} = (\mathcal{V}, \mathcal{E})$ denote a graph, where $\mathcal{V}$ and $\mathcal{E} \subseteq \mathcal{V} \times \mathcal{V}$ indicate the set of nodes and edges, respectively.
$\mathcal{G}$ is associated with the feature matrix $\mathrm{\textbf{X}}\in\mathbb{R}^{|{\mathcal{V}}|\times F}$ and the adjacency matrix $\mathbf{A} \in \mathbb{R}^{|\mathcal{V}| \times |\mathcal{V}|}$, where $F$ is the number of features, $\mathbf{A}_{ij} = 1$ if and only if $(v_i, v_j) \in \mathcal{E}$ and $\mathbf{A}_{ij} = 0$ otherwise.
Moreover, the nodes $\mathcal{V}$ are categorized into $\mathcal{C}$ classes, which are composed of base classes $\mathcal{C}_b$ and novel classes $\mathcal{C}_n$, i.e., $\mathcal{C}_b \cup \mathcal{C}_n = \mathcal{C} $ and $\mathcal{C}_b \cap \mathcal{C}_n = \varnothing $.

\subsubsection*{\textbf{Task: N-way K-shot Few-shot Node Classification}}
Given a graph $\mathcal{G}$ along with $\mathbf{X}$ and $\mathbf{A}$, we aim to learn a classifier $f(\cdot)$, which is able to generalize to new classes in meta-test tasks $\mathcal{T}^{te}$ by transferring meta knowledge obtained from meta-train tasks $\mathcal{T}^{tr}$.
Specifically, each meta-train task $\mathcal{T}_i^{tr}$ consists of support set $\mathcal{S}_i$ and query set $\mathcal{Q}_i$, whose nodes belong to $N$ classes that are sampled from the base classes $\mathcal{C}_b$.
Given randomly sampled $K$ support nodes and $M$ query nodes for each class, i.e., $N \times K$ nodes and $N \times M$ nodes, $f(\cdot)$ learns to classify the query node into given classes {based solely on the information provided by the support set}.
Then, our model generalizes to meta-test task $\mathcal{T}_i^{te}$, whose support set $\mathcal{S}_i$ and query set $\mathcal{Q}_i$ consist of $N$ novel classes in $\mathcal{C}_n$.

\section{Methodology}

\begin{figure*}[t] 
\begin{center}
\includegraphics[width=1.0\linewidth]{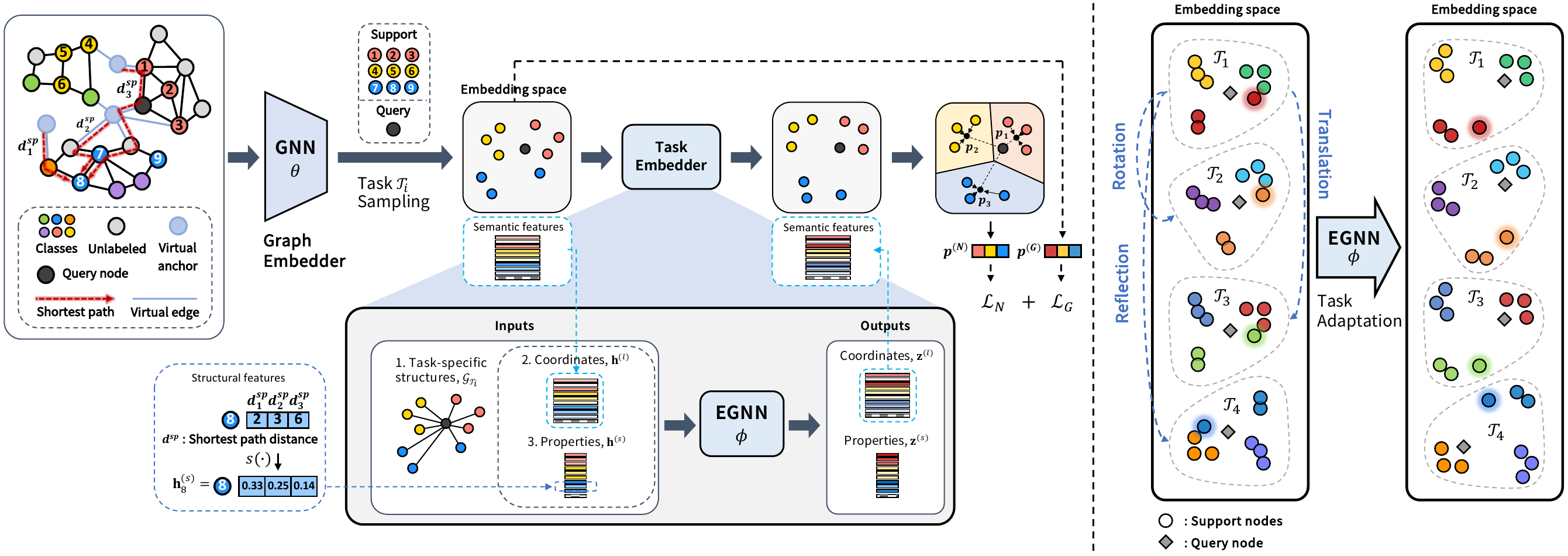}
\end{center}
\caption{Overview of \proposed~ in a 3-way 3-shot 1-query (N-way K-shot M-query) case (\textbf{left}). Our task embedder is equivariant to the following 3 types on a set of node representations $\mathbf{\textbf{h}}^{(l)}$ in a meta-task. \textit{(Rotation and reflection)} : For any orthogonal matrix $\textbf{Q}\in\mathbb{R}^{d_l\times d_l}$, $\mathbf{\textbf{h}}^{(l)}_{\mathcal{T}_2 (or \mathcal{T}_4)}=\mathbf{\textbf{h}}^{(l)}_{\mathcal{T}_1}\textbf{Q}$, then task adaptation strategy can be shared as $\phi(\mathbf{\textbf{h}}^{(l)}_{\mathcal{T}_1}\mathbf{Q})=\mathbf{\textbf{z}}^{(l)}_{\mathcal{T}_1}\mathbf{Q}=\mathbf{\textbf{z}}^{(l)}_{\mathcal{T}_2 (or \mathcal{T}_4)}=\phi(\mathbf{h}^{(l)}_{\mathcal{T}_2 (or \mathcal{T}_4)})$. \textit{(Translation)} : For any translation matrix $\textbf{G}\in\mathbb{R}^{|\mathcal{T}|\times d_l}$, $\mathbf{\textbf{h}}^{(l)}_{\mathcal{T}_3}=\mathbf{\textbf{h}}^{(l)}_{\mathcal{T}_1}+\textbf{G}$, then $\phi(\mathbf{\textbf{h}}^{(l)}_{\mathcal{T}_1}+\textbf{G})=\mathbf{\textbf{z}}^{(l)}_{\mathcal{T}_1}+\textbf{G}=\mathbf{\textbf{z}}^{(l)}_{\mathcal{T}_3}=\phi(\mathbf{\textbf{h}}^{(l)}_{\mathcal{T}_3})$ (right)}
\label{fig:overall}
\vspace{-1ex}    
\end{figure*}

In this section, we propose our method, called~\proposed, which follows the episodic training paradigm of meta-learning.
In a nutshell, our model learns task-specific embeddings for each node that are equivariant to transformations (e.g., rotation, reflection, and translation) in the representation space, enabling faster adaptation to the tasks of similar \textit{task-patterns}. Figure~\ref{fig:overall} describes the overview of~\proposed.

\subsection{Generating General Features}
Before computing the task-specific embeddings for each node, we generate two general representations of the nodes, i.e., semantic feature $\mathbf{H}^{(l)}\in \mathbb{R}^{|\mathcal{V}|\times d_l}$ and structural feature $\mathbf{H}^{(s)}\in \mathbb{R}^{|\mathcal{V}| \times d_s}$, which will be used as inputs of the task embedder, where $d_l$ and $d_s$ are the dimension size of the semantic feature and the structural feature, respectively.
Each feature captures the node's attributes and its structural location in the graphs, respectively, allowing for the classification of nodes based on both the semantic and structural contexts.

\subsubsection*{\textbf{Learning Semantic Feature}}
We learn the semantic feature of the nodes by mapping the nodes into a representation space via graph neural networks (GNNs), reflecting the semantic context of the entire graph through node attributes.
Specifically, we employ GCNs \cite{gcn} as a graph encoder to obtain the semantic feature $\mathbf{H}^{(l)}$ as follows:
\begin{equation}
    \textrm{\textbf{H}}^{(l)} = \textrm{GNN}_\theta(\textrm{\textbf{X},\textbf{A}}),
\end{equation}
where $\theta$ is the parameters of GNNs, and $\mathbf{X}$ and $\mathbf{A}$ denote the feature matrix and adjacency matrix, respectively.
The semantic feature represents the coordinates of each node in the representation space for the task embedder.
With the given coordinate information, our task-equivariant embedder quickly adapts to the new tasks by utilizing the relative difference between the coordinates of the nodes, even if the features are transformed in the representation space.
We will provide more detail {regarding how our task embedder adapts to the given meta-task} in Section \ref{taskembedder}.

\subsubsection*{\textbf{Generating Structural Features}}
However, relative differences in semantic features between nodes within a meta-task vary depending on the meta-task the node belongs to.
That is, even if the same node is involved in different meta-tasks, the model may not recognize them as the same node.
Therefore, it is crucial to utilize the structural feature such as the shortest path distance between nodes, which remains constant across all meta-tasks, since such structural features enable the model to identify the node regardless of the meta-tasks, which makes it easier for the model to align across different meta-tasks \cite{rale,pgnn}.
However, due to the complex nature of real-world graphs, it is not trivial to assess the overall structural context of the entire graph.
For example, there may exist connected components in a graph where edges are not interconnected, and thus calculating the shortest path distance between nodes that reside in different connected components is infeasible.

To this end, we propose to generate the structural features $\mathbf{H}^{(s)}$, by calculating the shortest path distance between the virtually created anchor nodes and the target node.
Specifically, we create a set of $k$ virtual anchor nodes, i.e., $\mathcal{V}_\alpha=\{v_{\alpha_{1}}, \ldots, v_{\alpha_{k}}\}$, each of which is connected to the nodes in the original graph with varying degrees of connectivity.
More precisely, we independently establish an edge between each virtual anchor node $v_{\alpha_{i}}\in\mathcal{V}_\alpha$ and node $v\in\mathcal{V}$ with a probability of $\frac{1}{2^i}$.
Building virtual anchor nodes with various degrees increases the certainty of structural information by connecting different components with high probability.
For example, virtual anchor nodes with small degrees provide highly accurate structural information as they are only connected to a few nearby nodes, while virtual anchor nodes with high degrees have a higher probability of connecting different components but provide less information about the structural context \cite{pgnn}.
Therefore, by generating virtual anchor nodes of various degrees of connectivity, the structural information can be maintained while sufficiently ensuring connectivity.
Then, we obtain structural features $\mathrm{\mathbf{H}}^{(s)}\in \mathbb{R}^{|\mathcal{V}| \times d_s}$ by measuring the shortest path distance between the target node and each anchor node.
More formally, we have:
\begin{equation}
    \mathrm{\mathbf{H}}^{(s)}_v=(s(v,v_{\alpha_1}),s(v,v_{\alpha_2}), \dots , s(v,v_{\alpha_k})),
\end{equation}
where $s(v,u)= 1/{(d^{sp}(v,u)+1)}$ and $d^{sp}(u,v)$ is the shortest path distance between node $v$ and $u$.
Note that once we generate the structural features, we can utilize them throughout the entire training and evaluation process. 

\subsection{Task-Equivariant Task Embedder} \label{taskembedder}
\subsubsection*{\textbf{Task Sampling}}
After obtaining the semantic feature $\mathbf{H}^{(l)}$ and structural feature $\mathbf{H}^{(s)}$ of the nodes in the entire graph, we sample tasks for episodic training.
In the case of $N$-way $K$-shot, in the meta-training phase, $K$ support nodes and $M$ query nodes are sampled for each class belonging to the $C_b$, i.e., $N\times (K+M)$ nodes for each task.
The only difference in the meta-test phase is to sample nodes belonging to $C_n$ instead of $C_b$.

\subsubsection*{\textbf{Task-equivariant Graph Neural Networks}}  
Now, we introduce the task embedder, which enforces \textit{task-equivariance} by utilizing Equivariant Graph Neural Networks (EGNN) \cite{egnn} as the encoder.
The inputs for our task embedder are the coordinates, i.e., semantic features, of each node in the embedding space denoted by $\mathrm{\mathbf{h}}^{(l)}$, and an embedding of the properties, i.e., structural features, of each instance denoted by $\mathrm{\mathbf{h}}^{(s)}$.
With the task embedder, we aim to update the coordinates via the relationship between support and query sets, regardless of their absolute coordinates that indicate their specific class.
By doing so, different from existing FSL methods on graphs, which fall short of generalizing well given a limited number of training meta-tasks with low diversity
by utilizing the absolute embeddings of the nodes, our approach extracts meta-knowledge by learning \textit{task-patterns}, which represent the relationships between nodes within meta-task.
This enables us to develop a highly transferrable task-adaptation strategy. 
Specifically, we first construct a task-specific graph structure $\mathcal{G}_{\mathcal{T}_i}$ that connects all pairs of support and query nodes.
Then, we generate a message $\mathrm{\mathbf{m}}_{ij}$ from node $j$ to $i$ with the properties of the nodes and the difference between the coordinates of the nodes as follows:
\begin{equation}
    \mathrm{\mathbf{m}}_{ij} = \phi_m(\mathrm{\mathbf{h}}^{(s),\lambda}_i, \mathrm{\mathbf{h}}^{(s),\lambda}_j, \lVert \mathrm{\mathbf{h}}^{(l),\lambda}_i-\mathrm{\mathbf{h}}^{(l),\lambda}_j \rVert ^2), \label{msggen}
\end{equation}
where $\mathrm{\mathbf{h}}^{(l),\lambda}$ and $\mathrm{\mathbf{h}}^{(s),\lambda}$ are coordinates and properties of the nodes at layer $\lambda$, respectively.
In Equation \ref{msggen}, we concatenate the properties of each node $\mathrm{\mathbf{h}}^{(s)}_i$,$\mathrm{\mathbf{h}}^{(s)}_j$ and the relative squared distance between two instances $\lVert \mathrm{\mathbf{h}}^{(l)}_i-\mathrm{\mathbf{h}}^{(l)}_j \rVert ^2$. 
Then, we pass them through an EGNN $\phi_m : \mathbb{R}^{2d_s+1} \rightarrow \mathbb{R}^{d_l}$.
With the generated messages $\mathbf{m}_{ij}$, we update coordinates and properties as follows:
\begin{equation}
\mathrm{\mathbf{h}}^{(l),\lambda+1}_i=\mathrm{\mathbf{h}}^{(l),\lambda}_i+{\frac{1}{C}}\sum_{j\neq i}(\mathrm{\mathbf{h}}^{(l),\lambda}_i-\mathrm{\mathbf{h}}^{(l),\lambda}_j)\phi_l(\mathrm{\mathbf{m}}_{ij}) \label{lttupt}\\
\end{equation}
\begin{equation}
\mathrm{\mathbf{m}}_{i}=\sum_{j \in \mathcal{N}(i)}\mathrm{\mathbf{m}}_{ij} \label{msgagg}\\
\end{equation}
\begin{equation}
\mathrm{\mathbf{h}}^{(s),\lambda+1}_i=\phi_s(\mathrm{\mathbf{h}}^{(s),\lambda}_i, 
\mathrm{\mathbf{m}}_{i}), \label{strupt}
\end{equation}
where $\phi_l$ and $\phi_s$ are parametererized MLP with non-linearities, and $\mathcal{N}(i)$ is the set neighbors of node $i$.
By learning from the difference between the coordinates of the nodes, the model understands the direction and distance between each pair of nodes and subsequently applies the appropriate task-adaptation strategy (updating the coordinates).

In Equation \ref{lttupt}, we update the coordinates of each node via computing the weighted sum of the differences between the coordinates of node $i$ and $j$, where the weight depends on the message that is passed through the function $\phi_l : \mathbb{R}^{d_l} \rightarrow \mathbb{R}$.
$C$ denotes the number of nodes within a meta-task, excluding node $i$.
After aggregating messages from neighboring nodes in Equation \ref{msgagg}, we finally update the structural features through $\phi_s : \mathbb{R}^{d_s+d_l} \rightarrow \mathbb{R}^{d_s}$ in Equation \ref{strupt}.
As a result, the output of the task embedder can be summarized as follow: $(\mathrm{\mathbf{z}}^{(l)}$,   $\mathrm{\mathbf{z}}^{(s)})=\phi({\mathrm{\mathbf{h}}^{(l)},\mathrm{\mathbf{h}}^{(s)}},\mathcal{G}_{\mathcal{T}_i})$. 
The proof of the model's \textit{task-equivariance} can be found in the Appendix \ref{apx:equi-proof}.

\subsection{Model Optimization}
Based on the learned semantic feature $\mathrm{\mathbf{h}}^{(l)}$ and task-specific feature $\mathrm{\mathbf{z}}^{(l)}$, we calculate two losses $\mathcal{L}_N$ and $\mathcal{L}_G$.
These losses incentivize the model to learn a fast adaptation strategy and to make the semantic features of different classes to be more separable, respectively.

\subsubsection*{\textbf{Network Loss ($\mathcal{L}_N$)}}
{After completing the adaptation for the given task, we calculate the final prediction logits using a metric that is invariant to transformations. 
This allows us to have consistent logits across different meta-tasks with the same \textit{task-patterns}.
In this work, we utilize ProtoNet \cite{protonet} that employs the squared Euclidean distance, which an invariant metric to transformations (e.g., rotations, reflections, and translations), for calculating the final logits.
Specifically,} we first calculate the average of the final coordinates, $\mathrm{\mathbf{z}}^{(l)}$, for the same class, as follows:
\begin{equation}
\mathrm{\mathbf{p}}^{(N)}_c = \frac{1}{K}\sum_{i=1}^K \mathrm{\mathbf{z}}^{(l)}_{c,i},
\end{equation}
where $\mathrm{\mathbf{z}}^{(l)}_{c,i}$ refers to the final coordinates of the $i$-th support node, which belongs to class $c$.
We then determine the class probability of query node $\mathrm{\mathbf{z}}^{(l)}_{qry}$ of class $c$ as:

\begin{equation} \label{latentprob}
p(c|\mathrm{\mathbf{z}}^{(l)}_{qry})=\frac{\mathrm{exp}(-d(\mathrm{\mathbf{z}}^{(l)}_{qry}, \mathrm{\mathbf{p}}^{(N)}_c))}{\sum_{c'=1}^{N}\mathrm{exp}(-d(\mathrm{\mathbf{z}}^{(l)}_{qry},\mathrm{\mathbf{p}}^{(N)}_{c'}))},
\end{equation} 
where $d(\cdot,\cdot)$ denotes the squared Euclidean distance. 
Finally, we classify the query node by finding the class with the highest probability, $y_{qry}=\text{argmax}_{c}p(c|\mathrm{\mathbf{z}}^{(l)}_{qry})$.

In order to maximize the probability $p(c|\mathrm{\mathbf{z}}^{(l)}_{qry})$ in Equation \ref{latentprob}, the loss $\mathcal{L}_N$ is defined as:
\begin{equation} \label{lossN}
\mathcal{L}_N=\sum^M_q\sum^N_c-\mathbb{I}(y_q=c)\mathrm{log}(p(c|\mathrm{\mathbf{z}}^{(l)}_{q})),
\end{equation}
where $y_q$ is the ground truth label of the $q$-th query node in $\mathcal{Q}_i$ and $\mathbb{I}(\cdot)$ is an indicator function.
Network loss is utilized to optimize the model in order to learn task-adaptation strategies for meta-tasks with certain \textit{task-patterns}.

\subsubsection*{\textbf{Graph Embedder Classification Loss ($\mathcal{L}_G$)}}
On the other hand, we calculate the prototype for each class using $\mathrm{\mathbf{h}}^{(l)}$, which is the representation of the graph embedder $\textrm{GNN}_\theta$, as follows: 
\begin{equation}
\mathrm{\mathbf{p}}^{(G)}_c = \frac{1}{K}\sum_{i=1}^K \mathrm{\mathbf{h}}^{(l)}_{c,i},
\end{equation}
where $\mathrm{\mathbf{h}}^{(l)}_{c,i}$ denotes the $i$-th support node's semantic feature generated by the graph embedder belonging to the class $c$. 

\begin{equation} \label{lgprob}
p(c|\mathrm{\mathbf{h}}^{(l)}_{qry})=\frac{\mathrm{exp}(-d(\mathrm{\mathbf{h}}^{(l)}_{qry}, \mathrm{\mathbf{p}}^{(G)}_c))}{\sum_{c'=1}^{N}\mathrm{exp}(-d(\mathrm{\mathbf{h}}^{(l)}_{qry},\mathrm{\mathbf{p}}^{(G)}_{c'}))},
\end{equation}
The graph embedder is directly trained to maximize the probability $p(c|\mathrm{\mathbf{h}}^{(l)}_{qry})$ to learn more distinct representations for different classes \cite{hgnn}. 
In order to maximize the probability $p(c|\mathrm{\mathbf{h}}^{(l)}_{qry})$, the loss $\mathcal{L}_G$ is defined as follows:
\begin{equation} \label{lossG}
\mathcal{L}_G=\sum^M_q\sum^N_c-\mathbb{I}(y_q=c)\mathrm{log}(p(c|\mathrm{\mathbf{h}}^{(l)}_{q})).
\end{equation}

\subsubsection*{\textbf{Loss Function}} As a result, the total meta-training loss is given as follows:
\begin{equation}
\mathcal{L}(\theta,\phi) = \mathcal{\gamma}\mathcal{L}_N + (1-\mathcal{\gamma})\mathcal{L}_G,
\end{equation}
where $\gamma$ is a tunable hyperparameter. 
The impact of adjusting the wof the hyperparameter is provided in Section \ref{ablation}.

\begin{table*}[] 
\caption{Model performance on various datasets under different few-shot settings (Accuracy).}
\resizebox{\textwidth}{!}
{
{
\renewcommand{\arraystretch}{1.25}
\begin{tabular}{c||cccccc||cccccc}
\toprule
Dataset                 & \multicolumn{6}{c||}{\textbf{Cora-full}}                                                                                                                         & \multicolumn{6}{c}{\textbf{Amazon Clothing}}                                                                                                                    \\ \toprule
Method                  & 5way 1shot          & 5way 3shot          & \multicolumn{1}{c|}{5way 5shot}          & 10way 1shot         & 10way 3shot         & 10way 5shot         & 5way 1shot          & 5way 3shot          & \multicolumn{1}{c|}{5way 5shot}          & 10way 1shot         & 10way 3shot         & 10way 5shot          \\ \hline\hline
MAML                & 24.74 ± 3.20               & 28.32 ± 1.83               & \multicolumn{1}{l|}{30.13 ± 4.33}          & 10.11 ± 0.49               & 10.98 ± 1.02               & 12.89 ± 1.78                & 45.60 ± 7.16               & 58.82 ± 5.52               & \multicolumn{1}{l|}{64.88 ± 1.89}          & 29.00 ± 1.86               & 39.52 ± 2.99               & 43.98 ± 2.27               \\ \hline
ProtoNet                & 31.47 ± 1.65               & 39.49 ± 1.46               & \multicolumn{1}{l|}{44.98 ± 1.08}          & 19.75 ± 0.71               & 28.16 ± 1.73               & 31.34 ± 0.91                & 42.37 ± 2.42               & 57.74 ± 1.09               & \multicolumn{1}{l|}{62.83 ± 3.10}          & 34.51 ± 2.13               & 49.16 ± 2.72               & 54.16 ± 1.62               \\ \hline
Meta-GNN                & 51.57 ± 2.83               & 58.10 ± 2.57               & \multicolumn{1}{l|}{62.66 ± 5.58}          & 29.20 ± 2.36               & 32.10 ± 4.60               & 41.36 ± 2.25                & 70.42 ± 1.66               & 76.72 ± 2.65               & \multicolumn{1}{l|}{76.27 ± 1.87}          & 51.05 ± 1.53               & 56.70 ± 2.22               & 57.54 ± 3.71               \\ \hline
G-Meta                  & 45.71 ± 1.97               & 54.64 ± 2.24               & \multicolumn{1}{l|}{58.68 ± 5.16}          & 32.90 ± 0.84               & 46.60 ± 0.62               & 51.58 ± 1.23                & 61.71 ± 1.67               & 67.94 ± 1.99               & \multicolumn{1}{l|}{73.28 ± 1.84}          & 50.33 ± 1.62               & 62.07 ± 1.12               & 67.23 ± 1.79               \\ \hline
GPN                     & 51.09 ± 3.55               & 63.78 ± 0.66               & \multicolumn{1}{l|}{65.89 ± 2.53}          & 40.24 ± 1.94               & 50.49 ± 2.34               & 53.75 ± 2.13                & 61.39 ± 1.97               & 73.42 ± 2.77               & \multicolumn{1}{l|}{76.40 ± 2.37}          & 51.32 ± 1.30               & 64.58 ± 3.04               & 69.03 ± 0.98               \\ \hline
TENT                    & 54.19 ± 2.23               & 65.20 ± 1.99               & \multicolumn{1}{l|}{68.77 ± 2.42}          & 37.72 ± 2.08               & 48.76 ± 1.95               & 53.95 ± 0.81                & 75.52 ± 1.06               & 85.21 ± 0.79               & \multicolumn{1}{l|}{87.15 ± 1.13}          & 60.70 ± 1.66               & 72.44 ± 1.81               & 77.53 ± 0.76               \\ \hline
\proposed                     & \textbf{60.27 ± 1.93}      & \textbf{74.24 ± 1.03}      & \multicolumn{1}{l|}{\textbf{76.37 ± 1.92}} & \textbf{45.26 ± 1.03}      & \textbf{60.00 ± 1.16}      & \textbf{64.56 ± 1.04}       & \textbf{80.77 ± 3.32}      & \textbf{90.14 ± 0.97}      & \multicolumn{1}{l|}{\textbf{90.18 ± 0.95}} & \textbf{69.12 ± 1.75}      & \textbf{79.42 ± 1.34}      & \textbf{83.27 ± 0.81}      \\ \bottomrule
\end{tabular} 
}
}

\resizebox{\textwidth}{!}
{
{
\renewcommand{\arraystretch}{1.25}
\begin{tabular}{c||cccccc||cccccc}
\toprule
Dataset                 & \multicolumn{6}{c||}{\textbf{Amazon Electronics}}                                                                                                                         & \multicolumn{6}{c}{\textbf{DBLP}}                                                                                                                    \\ \toprule
Method                  & 5way 1shot          & 5way 3shot          & \multicolumn{1}{c|}{5way 5shot}          & 10way 1shot         & 10way 3shot         & 10way 5shot         & 5way 1shot          & 5way 3shot          & \multicolumn{1}{c|}{5way 5shot}          & 10way 1shot         & 10way 3shot         & 10way 5shot          \\ \hline\hline
MAML                & 41.57 ± 6.32               & 54.88 ± 2.84               & \multicolumn{1}{l|}{62.90 ± 3.81}          & 28.75 ± 1.70               & 40.75 ± 3.20               & 41.98 ± 5.38                & 31.57 ± 3.57               & 43.52 ± 5.50               & \multicolumn{1}{l|}{51.09 ± 5.68}          & 16.05 ± 2.27               & 25.64 ± 2.24               & 25.66 ± 5.12               \\ \hline
ProtoNet                & 42.38 ± 1.62               & 52.94 ± 1.31               & \multicolumn{1}{l|}{59.34 ± 2.06}          & 32.05 ± 3.23               & 43.26 ± 1.72               & 49.49 ± 3.01                & 35.12 ± 0.95               & 49.27 ± 2.70               & \multicolumn{1}{l|}{53.65 ± 1.62}          & 24.30 ± 0.76               & 39.42 ± 2.03               & 44.06 ± 1.57               \\ \hline
Meta-GNN                & 57.23 ± 1.54               & 66.19 ± 2.40               & \multicolumn{1}{l|}{70.08 ± 2.14}          & 41.22 ± 2.85               & 48.94 ± 1.87               & 53.55 ± 1.51                & 63.07 ± 1.49               & 71.76 ± 2.17               & \multicolumn{1}{l|}{74.70 ± 2.09}          & 45.74 ± 1.68               & 53.34 ± 2.58               & 56.14 ± 0.88               \\ \hline
G-Meta                  & 47.14 ± 1.24               & 59.75 ± 1.29               & \multicolumn{1}{l|}{62.06 ± 1.98}          & 41.22 ± 1.86               & 48.64 ± 1.80               & 54.49 ± 2.37                & 57.98 ± 1.98               & 68.19 ± 1.40               & \multicolumn{1}{l|}{73.11 ± 0.81}          & 47.38 ± 2.72               & 60.83 ± 1.35               & 66.12 ± 1.79               \\ \hline
GPN                     & 48.32 ± 3.40               & 63.41 ± 1.54               & \multicolumn{1}{l|}{68.48 ± 2.38}          & 40.34 ± 1.86               & 53.82 ± 1.24               & 59.58 ± 1.39                & 60.43 ± 3.06               & 68.90 ± 0.54               & \multicolumn{1}{l|}{74.03 ± 1.77}          & 49.73 ± 1.64               & 62.34 ± 1.67               & 64.48 ± 2.43               \\ \hline
TENT                    & 69.26 ± 1.32               & 79.12 ± 0.97               & \multicolumn{1}{l|}{81.65 ± 1.31}          & 56.93 ± 1.65               & 68.56 ± 2.05               & 72.72 ± 0.78                & 72.19 ± 1.92               & 81.84 ± 1.82               & \multicolumn{1}{l|}{82.76 ± 1.29}          & 58.40 ± 1.41               & 68.55 ± 1.38               & 72.47 ± 1.27               \\ \hline
\proposed                     & \textbf{73.78 ± 0.93}      & \textbf{84.78 ± 1.52}      & \multicolumn{1}{l|}{\textbf{87.17 ± 1.15}} & \textbf{61.34 ± 1.58}      & \textbf{76.48 ± 1.36}      & \textbf{79.63 ± 0.73}       & \textbf{74.32 ± 1.66}      & \textbf{83.10 ± 2.01}      & \multicolumn{1}{l|}{\textbf{83.33 ± 1.22}} & \textbf{61.81 ± 2.02}      & \textbf{71.25 ± 1.23}      & \textbf{74.50 ± 1.49}      \\ \bottomrule
\end{tabular}
}
\label{table:maintable}
}
\end{table*}

\section{EXPERIMENTS}
{In this section, we provide experimental settings (Sec.~\ref{setup}), present results demonstrating the high generalization power of our proposed framework (Sec.~\ref{sec:mainresults}), analyze the effectiveness of \textit{task-equivariance} (Sec.~\ref{sec:equi-proof}), verify the impact of the diversity of the meta-train tasks  and \proposed's ability to address low diversity (Sec.~\ref{sec:diversity}), and analyze the impact of components in \proposed~(Sec.~\ref{ablation}).}

\subsection{Experimental Setup} \label{setup}
\subsubsection*{\textbf{Datasets}}
{
We use five real-world datasets, i.e., \textbf{Cora-full} \cite{corafull}, \textbf{Amazon Clothing} \cite{amazon}, \textbf{Amazon Electronics} \cite{amazon}, \textbf{DBLP} \cite{dblp} and \textbf{Coauthor-CS} \cite{coauthor}, to comprehensively evaluate the performance of \proposed, whose details are provided in Appendix~\ref{app:datasets}.
We also provide statistics for the datasets in Table~\ref{table:datastats}.
}

\vspace{-0.5em}


\begin{table}[H] 
\normalsize
\centering
\caption{Statistics of evaluation datasets.}
\vspace{-3mm}
{
\renewcommand{\arraystretch}{0.9}
\resizebox{0.99\columnwidth}{!}{
\begin{tabular}{c|ccccc}
Dataset            & \# Nodes & \# Edges & \# Features  & Class Split    \\ \hline\hline
Cora-full           & 19,793   & 65,311   & 8,710                & 25/20/25                       \\ 
Amazon Clothing    & 24,919   & 91,680   & 9,034                & 40/17/20                        \\ 
Amazon Electronics & 42,318   & 43,556   & 8,669               & 91/36/40                        \\ 
DBLP               & 40,672   & 288,270  & 7,202               & 80/27/30                        \\ 
Coauthor-CS         & 18,333   & 163,788  & 6,805                & 5/5/5                            \\ \hline
\end{tabular}
}
\label{table:datastats}
}\end{table}

\vspace{-1em}

\subsubsection*{\textbf{Compared Methods}}
{
We compare \proposed~with six recent state-of-the-art methods, i.e., \textbf{MAML} \cite{maml}, \textbf{ProtoNet} \cite{protonet}, \textbf{Meta-GNN} \cite{metagnn},  \textbf{GPN} \cite{gpn}, \textbf{G-Meta} \cite{gmeta} and  \textbf{TENT} \cite{tent}, whose descriptions and implementation details can be found in Appendix~\ref{app:baselines}.
In brief, MAML and ProtoNet are representative methods for few-shot learning in the computer vision domain, while the other methods specifically target few-shot learning tasks on attributed graphs.
}

\subsubsection*{\textbf{Evaluation Protocol}}
We evaluate the performance of all the methods on six different settings of few-shot node classification tasks ($N$-way $K$-shot), i.e., 5/10-way 1/3/5-shot.
Each model is evaluated using 50 randomly sampled meta-testing tasks for a fair comparison.
We use a query size of 5 in our experiments, and all methods are trained using the same 500 randomly selected meta-tasks from the training dataset.
It is a common setting on FSL, which means all training data are available so that we can make high-diversity meta-training tasks. 
We measure the performance in terms of accuracy (ACC), which is averaged over 5 independent runs with different random seeds.
Further implementation details of \proposed~are provided in Appendix \ref{app: Implementation details}.

\subsection{Main Results} \label{sec:mainresults}

\begin{figure}[t] 
\begin{center}
\includegraphics[width=1.0\linewidth]{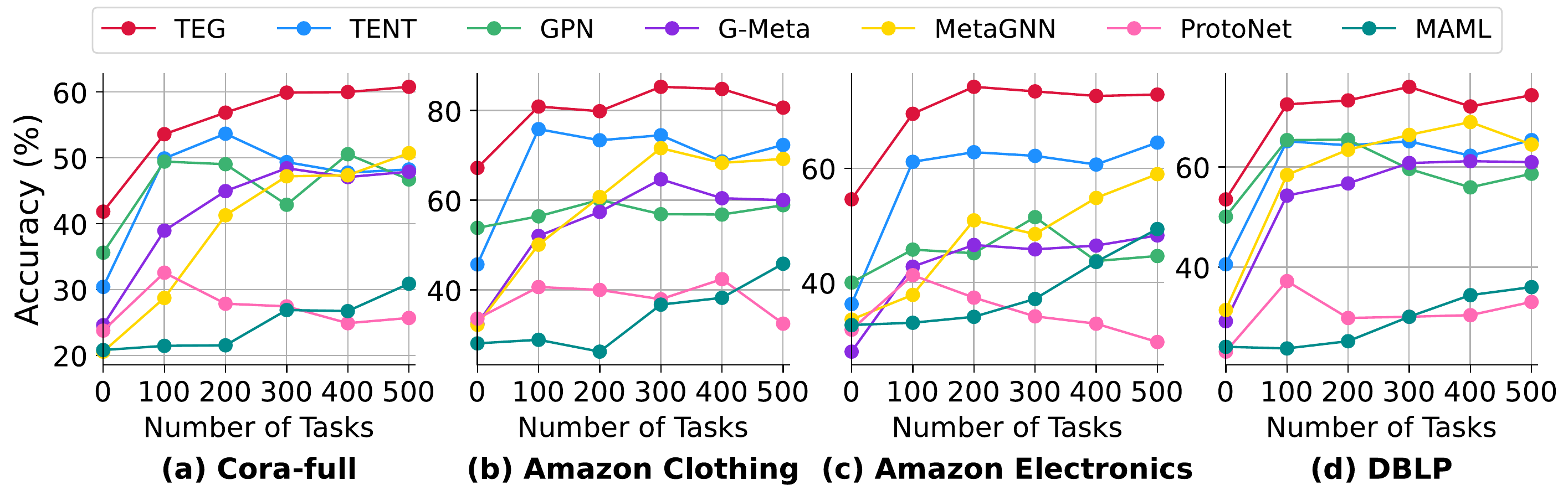}
\end{center}
\caption{Model performance on different datasets in 5-way 1-shot setting. \textit{Number of Tasks} denotes the number of trained meta-training tasks used during training as epochs increase.}
\vspace{-3mm}
\label{fig:curves}
\end{figure}

The experimental results on four datasets on six distinctive settings are summarized in Table \ref{table:maintable}.
We have the following observations:
\textbf{1)}
In general, \proposed~outperforms all the baselines that do not take the \textit{task-equivariance} in embedding space into account.
We attribute this to the generalization ability of \proposed, which is achieved by introducing the concept of equivariance into the model.
\textit{Task-equivariance} facilitates the adaptation of our model to unseen meta-tasks that have similar \textit{task-patterns} by enabling the sharing of the task-adaptation strategy.
A more detailed and intuitive analysis will be discussed in the following section.
\textbf{2)} 
Generally, increasing $K$ (i.e., the number of support nodes within each meta-task) improves the performance of all methods, as FSL relies on limited data.
FSL tends to be sensitive to reference data (i.e., support set), so having more references for each class leads to a more accurate representation, reducing the impact of outliers.
\textbf{3)} 
As the number of classes within each meta-task $N$ increases, the overall model performance deteriorates significantly since greater challenges in generalization are posed.
However, \proposed, which leverages instance relationships to extract meta-knowledge, still outperforms the baseline methods, again demonstrating the generalization ability of the model.
Specifically, the performance gap between our model and the runner-up grows larger as $N$ increases across all settings.
\textbf{4)} 
Additionally, as depicted in Figure \ref{fig:curves}, our model consistently outperforms baseline methods throughout all training steps.
Even in the initial training epochs, where all the models are trained with a limited number of meta-tasks, \proposed~shows a large performance improvement compared to the baseline methods.
This is because \proposed~ quickly extracts well-generalized meta-knowledge even with a small number of meta-tasks by utilizing the equivariant task embeddings.
\textbf{5)} 
It is worth noting that among the metric-based methods, the method with sophisticated task-specific embedding mechanisms to manage high variance in meta-tasks, i.e., TENT, outperforms the method with a simple graph embedder, demonstrating the importance of tasks-specific embedder.
However, we find out that TENT's performance decreases after reaching a peak in the early epochs as shown in Figure \ref{fig:curves} (a) and (b).
This is because TENT's approach of treating each meta-task independently results in reduced transferability of task-adaptation strategies, easily causing overfitting to the training class.
On the other hand, our task embedder is equipped with \textit{task-equivariance}, allowing it to learn transferable meta-knowledge and prevent overfitting to meta-training tasks.

To summarize our findings, \proposed~achieves great generalization power by utilizing \textit{task-equivariance} of the task embedder, even with minimal meta-training tasks.
We further demonstrate the generalization ability of \proposed~by probing the \textit{task-equivariance} of the task embedder in Section \ref{sec:equi-proof} and conducting experiments on the various levels of diversity in Section \ref{sec:diversity}.

\vspace{-1mm}



\subsection{Effectiveness of \textit{Task-Equivariance}} \label{sec:equi-proof}

\begin{figure}[t] 
    \begin{center}
    \includegraphics[width=1.0\linewidth]{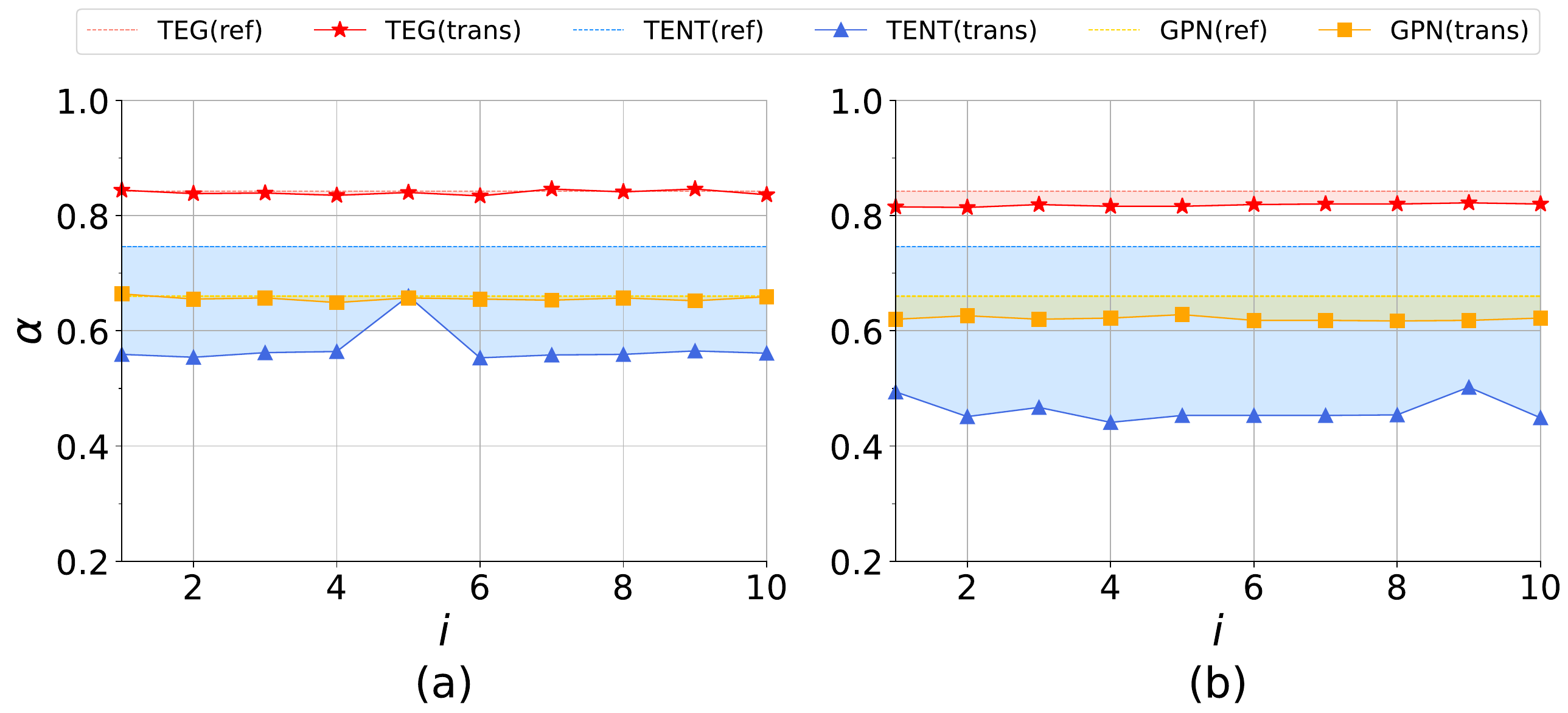}
    \end{center}
    \vspace{-1ex}
    \caption{Accuracy comparison on the (a) task transformation and (b) task transformation with noise injection.
    (Dotted line: model accuracy in $\mathrm{\textbf{T}}_{ref}$ /
    Solid line: model accuracy in $\mathrm{\textbf{T}}_{trans}$ /
    Filled color : performance gap between $\mathrm{\textbf{T}}_{ref}$ and $\mathrm{\textbf{T}}_{trans}$.)}
    \label{fig:equitest}
\vspace{-1ex}
\end{figure}

In order to verify the generalization ability of \proposed~achieved by the \textit{task-equivariance},
we evaluate the model performance on a set of meta-tasks generated by transforming the meta-train tasks set, and compare it to the performance of other metric-based methods, i.e., TENT \cite{tent} and GPN \cite{gpn}, in Figure \ref{fig:equitest} (a).
Specifically, we train all the methods with 500 meta-training tasks sampled from the Amazon-clothing dataset, referred to as the reference set $\mathrm{\textbf{T}}_{ref}=\{\mathcal{T}_1,\ldots,\mathcal{T}_{500}\}$, in a 10-way 5-shot setting, and $|\mathcal{T}|$ denotes the number of instances in a task $\mathcal{T}$.
Then, we generate a set of transformed tasks $\mathrm{\textbf{T}}_{trans}$ from $\mathrm{\textbf{T}}_{ref}$ by utilizing orthogonal matrices $\mathbf{Q}\in\mathbb{R}^{d\times d}$ for rotations and reflections, and translation matrices $\textbf{G}=\lambda\textbf{1}_{|\mathcal{T}|}\textbf{1}^\mathrm{T}_d\in\mathbb{R}^{|\mathcal{T}|\times d}$, with a randomly sampled scalar value $\lambda$, for translations.
That is, we pair the random orthogonal matrices and translation matrices to generate a transformation set of 10 different pairs, $S=\{{(\mathbf{Q}_1,\textbf{G}_1), (\mathbf{Q}_2,\textbf{G}_2),\dots, (\mathbf{Q}_{10},\mathbf{G}_{10})}\}$, and generate a set of transformed tasks $\mathrm{\textbf{T}}_{trans} = \{ \mathrm{\textbf{T}}_{ref}\mathbf{Q}_1+\textbf{G}_1, \ldots, \mathrm{\textbf{T}}_{ref}\mathbf{Q}_{10}+\textbf{G}_{10}\}$\footnote{Let $\mathrm{\textbf{T}}_{ref}\mathbf{Q}$ stand for $(\mathrm{\textbf{h}}^{(l)}_{\mathcal{T}_1}\mathbf{Q},\dots, \mathrm{\textbf{h}}^{(l)}_{\mathcal{T}_{500}}\mathbf{Q})$, and $\mathrm{\textbf{T}}_{ref}+\textbf{G}$ represent $(\mathrm{\textbf{h}}^{(l)}_{\mathcal{T}_1}+\textbf{G}[1,:],\dots, \mathrm{\textbf{h}}^{(l)}_{\mathcal{T}_{500}}+\textbf{G}[500,:])$.}. 
Then we compare the accuracy of each model in $\mathrm{\textbf{T}}_{ref}$ to that of $\mathrm{\textbf{T}}_{trans}$, which is measured by averaging the performance of each transformed task, i.e., $\alpha = acc(\mathrm{\textbf{T}}_{ref}\mathbf{Q}_i+\textbf{G}_i)$.
Moreover, to figure out how the models learn meta-knowledge when the meta-tasks have similar patterns but are not exactly identical to the transformation, we evaluate the accuracy of the transformed set after adding Gaussian noise to the node embeddings in Figure \ref{fig:equitest} (b).


Based on the experimental results, we have the following observations:
\textbf{1)} In Figure \ref{fig:equitest} (a), our model maintained almost the same level of performance as the original performance for the transformed meta-tasks\footnote{We expect that slight gap is caused by the process of generating or multiplying a high-dimensional orthogonal matrix.}.
It indicates that \proposed~utilizes the same adaptation strategy for meta-tasks with the same \textit{task-patterns} as meta-training tasks, verifying the proper functioning of its \textit{task-equivariance}.
\textbf{2)} Moreover, our model shows its robustness even if the noise is injected to the transformed tasks in Figure \ref{fig:equitest} (b).
This also demonstrates our model's generalizability, enabling effective adaptation to not only the transformed set of meta-tasks but also the meta-task with similar patterns.
\textbf{3)} 
On the other hand, since GPN only utilizes invariant features (e.g., raw features and node degrees) on graphs and does not implement extra task-adaptation, it still maintains a similar level of performance on the transformed tasks $\mathrm{\textbf{T}}_{trans}$ compared to that of reference tasks $\mathrm{\textbf{T}}_{ref}$.
However, such invariance results in poor performance on the reference tasks $\mathrm{\textbf{T}}_{ref}$ since it neglects the relationships between nodes within a meta-task during training.
\textbf{4)} Moreover, we observe that TENT cannot properly adapt to the transformed set because it doesn't have a \textit{task-equivariance} property.
Therefore, we argue the fragility of TENT, whose performance severely deteriorates with the subtle variance of tasks induced by transformation and noise injection despite its relatively competitive performance on the reference set.

In conclusion, our findings demonstrate that \textit{task-equivariance} enables the model to acquire highly transferable meta-knowledge that can be applied to new tasks with both same and similar \textit{task-patterns}.

\vspace{-0.5em}

\begin{figure*}[h]
  \includegraphics[width=0.92\textwidth]{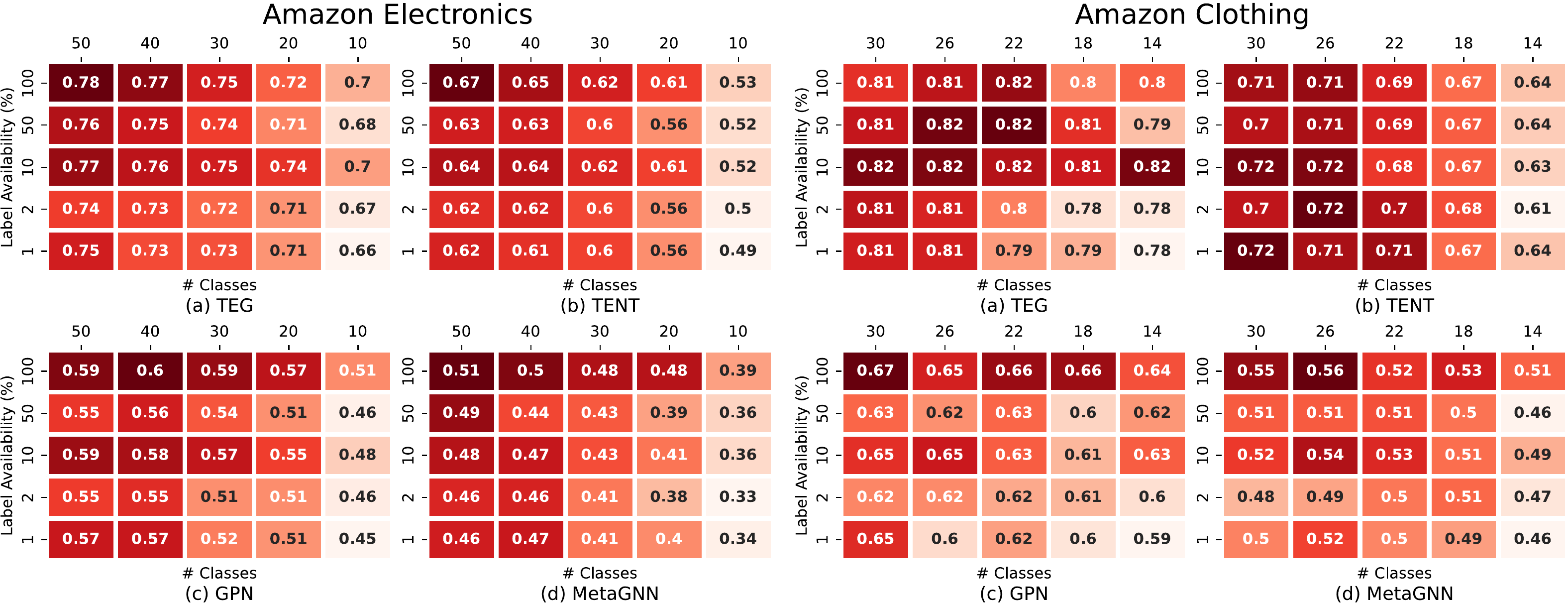}
  \vspace{-2ex}
  \caption{Impact of diversity in 10way-5shot meta-training tasks. The test accuracy is written in the box with the color indicating the level. \textit{Label availability} denotes the percentage of labeled train data that is available to use for each class, and \textit{\# Classes} denotes the number of train classes. Meta-tasks' diversity decreases as it goes toward the bottom right of the heatmap.}
  \label{fig:heatmap}
\vspace{-1ex}
\end{figure*}

\begin{table*}[h] 
\caption{Model performance on the limited diversity of meta-training dataset (Accuracy).} 
\vspace{-1ex}
\label{table:clslbl}
\resizebox{\textwidth}{!}
{
\renewcommand{\arraystretch}{0.82}
\begin{tabular}{c||cccccc||cccccc}
\toprule
Dataset            & \multicolumn{6}{c||}{\textbf{Amazon Electronics}}                                                 & \multicolumn{6}{c}{\textbf{Amazon Clothing}}                                                    \\ \hline
Setting            & \multicolumn{3}{c|}{5way 5shot}                      & \multicolumn{3}{c||}{10way 5shot} & \multicolumn{3}{c|}{5way 5shot}                      & \multicolumn{3}{c}{10way 5shot} \\ \hline
Class/label Avail. & 50\%/10\% & 30\%/2\% & \multicolumn{1}{c|}{10\%/1\%} & 50\%/10\%  & 30\%/2\% & 10\%/1\% & 50\%/10\% & 30\%/2\% & \multicolumn{1}{c|}{10\%/1\%} & 50\%/10\% & 30\%/2\% & 10\%/1\% \\ \hline\hline
MAML            & 58.50     & 55.10    & \multicolumn{1}{c|}{52.00}    & 44.31      & 40.48    & 34.04    & 58.62     & 53.30    & \multicolumn{1}{c|}{50.16}    & 38.22     & 33.70    & 34.46    \\ \hline
ProtoNet            & 54.93     & 54.86    & \multicolumn{1}{c|}{47.15}    & 47.75      & 42.80    & 33.93    & 57.78     & 51.89    & \multicolumn{1}{c|}{46.74}    & 43.21     & 37.22    & 37.02    \\ \hline
Meta-GNN            & 68.10     & 62.45    & \multicolumn{1}{c|}{56.24}    & 47.70      & 41.23    & 33.86    & 75.28     & 73.73    & \multicolumn{1}{c|}{66.29}    & 54.18     & 50.83    & 45.70    \\ \hline
G-Meta                & 58.62     & 53.30    & \multicolumn{1}{c|}{50.16}    & 38.22      & 33.70    & 34.46    & 58.50     & 55.10    & \multicolumn{1}{c|}{52.00}    & 44.31     & 40.48    & 34.04    \\ \hline
GPN                & 69.68     & 62.14    & \multicolumn{1}{c|}{55.33}    & 58.66      & 51.06    & 45.51    & 73.06     & 71.06    & \multicolumn{1}{c|}{70.66}    & 65.25     & 61.24    & 60.59    \\ \hline
TENT               & 74.90     & 70.66    & \multicolumn{1}{c|}{56.16}    & 64.43      & 60.11    & 48.46    & 80.40     & 77.38    & \multicolumn{1}{c|}{65.15}    & 68.91     & 63.16    & 60.46    \\ \hline
\proposed                & \textbf{83.26}     & \textbf{81.84}    & \multicolumn{1}{c|}{\textbf{76.77}}    & \textbf{75.37}      & \textbf{72.61}    & \textbf{68.98}    & \textbf{88.26}     & \textbf{86.72}    & \multicolumn{1}{c|}{\textbf{82.54}}    & \textbf{80.88}     & \textbf{78.76}    & \textbf{78.41}    \\ \hline\hline
Rel Improv.        & 11.2\%    & 15.8\%   & \multicolumn{1}{c|}{36.5\%}   & 17.0\%     & 20.8\%   & 42.3\%   & 9.8\%     & 12.1\%   & \multicolumn{1}{c|}{16.8\%}   & 17.4\%    & 24.7\%   & 29.4\%   \\ \bottomrule
\end{tabular}
}
\end{table*}

\subsection{Impact of Diversity of Meta-Train Tasks} \label{sec:diversity}

In this section, we examine how the diversity of meta-train tasks affects the performance of models using an episodic meta-learning approach by conducting experiments on a limited number of classes and data instances.
Specifically, we train the models with the training set that contains the various portion of the whole classes and instances in each class, then evaluate them on meta-testing tasks containing all classes and instances in each class.
The overall results are depicted in Figure \ref{fig:heatmap} and Table \ref{table:clslbl}.
We have the following observations:
\textbf{1)} In Figure \ref{fig:heatmap}, the performance of all models degrades as the number of classes and label availability decrease (i.e., towards bottom right corner).
On the other hand, we observe that when either the number of classes or label availability is sufficient, models do not suffer from performance degradation.
This indicates sufficient information of either one makes up for the lack of diversity of the counterpart, thereby maintaining the overall performance.
\textbf{2)} Meanwhile, we observe that the performance of \proposed~in an extreme case, i.e., 10 classes with 1\% instances each, is comparable to that of other methods in the easiest case, i.e., 50 classes with 100\% instances each.
This highlights our model's high generalization ability to obtain meta-knowledge for a wide range of meta-tasks with limited training meta-tasks.
\textbf{3)} Moreover, our model achieves further performance improvements compared to the baseline methods as the diversity of tasks decreases as shown in Table \ref{table:clslbl} in which the diversity of tasks decreases from 50\%/10\% to 10\%/1\% in each setting.
This result demonstrates the real-world applicability of our model, where an abundant amount of labels and data are not provided.
To summarize our findings, \proposed~ outperforms other models when faced with limited meta-training tasks and has a strong ability to adapt to new tasks with minimal training data, which is common in real-world scenarios.

\begin{table}[t] 
\caption{Model performance on Coauthor-CS dataset under different few-shot settings (Accuracy).} 
\vspace{-1ex}
\label{table:coauthor}
\resizebox{\columnwidth}{!}
{
{
\renewcommand{\arraystretch}{0.83}
\begin{tabular}{c||cccccc}
\toprule
Dataset & \multicolumn{6}{c}{\textbf{Coauthor-CS}}             \\ \hline
Setting            & \multicolumn{3}{c|}{2-way}     &  \multicolumn{3}{c}{5-way}    \\ \hline
Method & 1shot & 3shot & \multicolumn{1}{c|}{5shot} & 1shot  & 3shot & 5shot \\ \hline\hline
MAML            & 67.92     & 78.60    & \multicolumn{1}{c|}{80.64}    & 40.05      & 49.98    & 49.47 \\ \hline
ProtoNet            & 65.52     & 69.84    & \multicolumn{1}{c|}{74.44}    & 40.21      & 52.98    & 51.71 \\ \hline
Meta-GNN            & 81.88     & 87.20    & \multicolumn{1}{c|}{89.04}    & 55.58      & 62.16    & 64.19 \\ \hline
G-Meta                & 65.25     & 77.95    & \multicolumn{1}{c|}{77.67}    & 52.97      & 63.83    & 67.65  \\ \hline
GPN                & 80.56     & 87.96    & \multicolumn{1}{c|}{88.80}    & 60.61      & 71.46    & 73.62  \\ \hline
TENT               & 88.12     & 90.28    & \multicolumn{1}{c|}{88.36}    & 57.18      & 71.70    & 74.14    \\ \hline
\proposed                & \textbf{92.40}     & \textbf{94.04}    & \multicolumn{1}{c|}{\textbf{95.36}}    & \textbf{75.07}      & \textbf{84.83}    & \textbf{83.70}  \\ \hline\hline
Rel Improv.               & 4.9\%     & 4.2\%    & \multicolumn{1}{c|}{7.1\%}    & 23.9\%      & 18.3\%    & 12.9\%    \\ \bottomrule
\end{tabular}
}
}
\vspace{-2.5ex}
\end{table}

We further investigate the impact of diversity with Coauthor-CS dataset, which is a real-world dataset inherently having a few classes.
Specifically, we fix train/valid/test classes into 5/5/5, respectively, and then compare the model performance in 2 main settings: 2-way $K$-shot and $5$-way $K$-shot settings.
Note that, in the 2-way settings, all methods can be trained with meta-training tasks composed of 10 class combinations (i.e., $\Comb{5}{2}$) ensuring an acceptable level of diversity within the meta-tasks.
In the 5-way settings, however, only one combination (i.e., $\Comb{5}{5}$) of classes can be used, making it challenging for the baselines to extract meta-knowledge due to the low diversity within the meta-tasks.
As shown in Table \ref{table:coauthor}, we find out that the performance improvements of~\proposed~compaerd with the best-performing baseline are significantly higher in the 5-way settings (i.e., low-diversity scenarios), demonstrating the outstanding ability to extract meta-knowledge from low-diversity meta-tasks.
Therefore, we argue that previous meta-learning baselines struggle to extract meta-knowledge when faced with the low-diversity meta-tasks, while our model effectively handles low-diversity scenarios with exceptional generalization ability.

We further proposed a novel strategy for constructing meta-tasks when there is a lack of training classes (See Appendix \ref{apx:experiments}).

\subsection{Model Analysis} \label{ablation}

In this section, we investigate the impact of various components in \proposed, i.e., the number of virtual anchor nodes $k$ and the hyperparameter $\gamma$.

\subsubsection*{\textbf{Impact of the Number of Virtual Anchor Nodes $k$}}

\begin{figure}[t] 
    \begin{center}
    \includegraphics[width=1.0\linewidth]{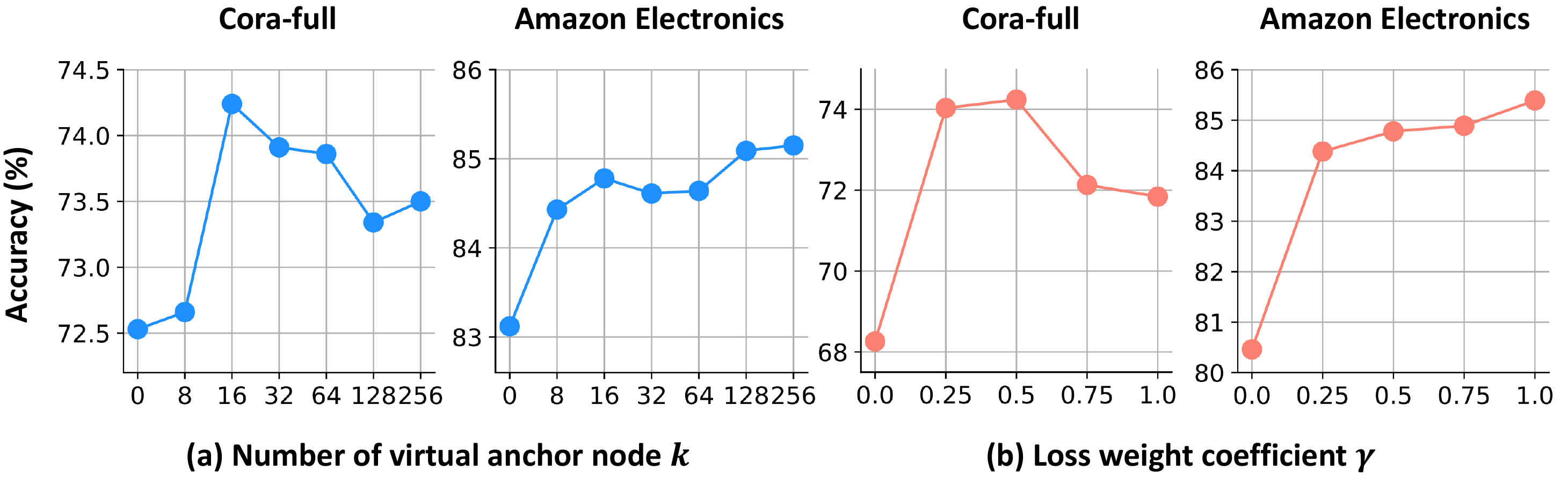}
    \end{center}
    \vspace{-2mm}
    
    \caption{Model analysis in 5way-3shot setting.}
    \label{fig:sensitivity}
\vspace{-2em}
\end{figure}

We investigate the impact of the different numbers of virtual anchor nodes $k$, which also defines the dimension of structural feature $\mathbf{H}^{(s)}\in \mathbb{R}^{|\mathcal{V}| \times k}$.
From Figure \ref{fig:sensitivity}, we have the following observations:
\textbf{1)} Using virtual nodes greatly improves model performance compared to that of not using virtual anchor nodes, i.e., $k = 0$.
This is because the virtual anchor nodes generate structural features that are invariant features of each node, making it easy for the model to align meta-tasks while providing auxiliary information regarding the node's position within the entire graph.
\textbf{2)} On the other hand, the correlation between the model performance and the number of virtual anchor nodes varies according to a characteristic of the dataset, i.e., sparsity.
More specifically, in Figure \ref{fig:sensitivity} (a), we observe the model performance increases as the number of virtual anchor nodes increases in the Amazon Electronics dataset, while it was not the case in the Cora-full dataset.
This is because the Amazon Electronics dataset is much sparser than Cora-full dataset, 
i.e., containing more connected components in a graph with no connections between them, 
which allows our model to fully benefit from utilizing more virtual anchor nodes.
However, adding too many virtual anchor nodes in Cora-full dataset, which has dense node connectivity, may distort the structure of the original graph, leading to a decrease in the model performance.
{In the Appendix~\ref{apx:effect_virtual}, we provide additional experiments for exploring the effect of using virtual anchor nodes.}

\subsubsection*{\textbf{Impact of hyperparameter $\gamma$}} \label{gamma}

In this section, we evaluate the relative importance between the two losses, i.e., $\mathcal{L}_{N}$ and $\mathcal{L}_{G}$, by varying the hyperparameter $\gamma$ from 0 to 1.
Note that when $\gamma = 0$, the graph embedder is solely trained without the task embedder, while the direct optimization of the graph embedder is excluded when $\gamma = 1.0$.
Based on Figure \ref{fig:sensitivity} (b), we have the following observations:
\textbf{1)} In general, model performance deteriorates severely when the task embedder is not utilized during training. 
This demonstrates the importance of the task embedder that learns the \textit{task-equivariance}, which is also the main contribution of our work.
\textbf{2)} However, the effect of the graph embedder loss $\mathcal{L}_{G}$ depends on the dataset.
Specifically, we observe that the model performs the best when both losses have the same effect during training in Cora-full dataset (i.e. $\gamma = 0.5$), while utilizing only the task embedding loss $\mathcal{L}_{N}$ is beneficial to Amazon Electronics dataset (i.e. $\gamma = 1.0$).
This indicates that the key to the success of \proposed~is the task embedder that learns the \textit{task-equivariance}, thereby effectively training not only the task embedder itself but also the graph embedder.


{
\section{Related Work}
\subsubsection*{\textbf{Few-shot Learning.}} The goal of few-shot learning (FSL) is to develop a model that is able to generalize to new classes, even when it has only been exposed to a limited number of examples, typically one or a few examples from those classes.
Meta-learning is a widely used approach for FSL as it possesses the capability to endow the model with robust generalization ability.
Specifically, it extracts meta-knowledge, which is common across various meta-tasks, allowing the model to adapt to new unseen meta-tasks by leveraging sufficient meta-training tasks. 
It can be classified into two main categories; \textit{metric-based approaches}, and \textit{optimization-based approaches}.
\textit{Metric-based approaches} \cite{protonet,sung2018learning, matchingnet,liu2019learning} learn a metric function that measures the similarity between new instances and a limited number of examples by mapping instances to a metric space.
For instance, Matching Network \cite{matchingnet} obtains representations of both the support and query sets by using separate encoders, to compute a similarity between them.
ProtoNet \cite{protonet} utilizes prototypes, which are obtained by averaging the representations of the support set for each class. 
It then classifies a query node by computing the Euclidean distance between the query node's representation and the prototypes of each class.
\textit{Optimization-based approaches} \cite{maml,li2017meta,nichol2018first,ravi2016optimization,ravi2017optimization} learn how to optimize the model parameters using the gradients from a few examples.
For example, MAML \cite{maml} optimizes the initial parameters where the model can be quickly fine-tuned with a small number of steps using a few examples. 
Another example is a meta-learner that utilizes an LSTM to learn appropriate parameter updates for a classifier specifically for a given task \cite{ravi2017optimization}.
Recently, FSL on graphs employ meta-learning based approaches \cite{gpn, tent, rale, metagnn, gmeta, metagps} and have been successful in their application to attributed networks.
Nevertheless, in order to extract meta-knowledge that can lead to satisfactory performance, current methods require a significant number of meta-training tasks that contain a diverse set of classes and the corresponding nodes.
Our framework, on the other hand, can alleviate the issue of limited diversity in meta-tasks.
}
\vspace{-0.5em}

\section{Conclusion \& Future Work}
In this work, we propose \proposed, a novel few-shot learning framework that learns highly transferable task-adaptation strategies.
The main idea is to learn the task-specific embeddings of the nodes that are equivariant to transformations (e.g., rotation, reflection and translation) by utilizing an equivariant neural network as a task embedder.
By doing so, \proposed~learns generalized meta-knowledge that can be applied to the tasks with similar \textit{task-patterns}, even if the classes and node representations within each meta-task are different.
We demonstrate the effectiveness of \textit{task-equivariance} by validating its generalization power across meta-tasks with shared \textit{task-patterns}.
Extensive experiments on various few-shot learning settings demonstrate the superiority of \proposed.
A further appeal of \proposed~is its generalization ability even with the limited diversity of data, which is common in real-world scenarios.  

\smallskip
\noindent\textbf{Future Work. }
In recent studies on FSL, besides meta-learning based methods, other works have also highlighted the importance of high-quality representations of data for solving FSL tasks \cite{tian2020rethinking, tan2022transductive}. 
Notably, it has been shown that self-supervised learning (SSL) methods, such as graph contrastive learning (GCL), outperform existing meta-learning-based methods \cite{tan2022transductive}.
This has led to the perspective that SSL methods may be more effective solutions for FSL compared to meta-learning.
However, while SSL methods \cite{zhu2020deep, thakoor2021bootstrapped, lee2022augmentation, lee2022relational} focus on learning high-quality representations of nodes/edges/graphs, meta-learning-based approaches focus on training a high-quality classifier with high generalization power. 
Recent research \cite{chen2021meta} has also shown that the choice of metric used for query classification can impact the performance of whole-classification models like GCL.
Given these insights, we propose exploring the combination of high-quality node embeddings derived from SSL with a carefully designed classifier obtained through meta-learning, as a potential avenue to further enhance model performance on FSL tasks.

\vspace{-0.5em}

\noindent \subsubsection*{\textbf{Acknowledgement.}}
This work was supported by Institute of Information \& communications Technology Planning \& Evaluation (IITP) grant funded by the Korea government(MSIT) (No.2022-0-00157 and No.2022-0-00077).

\bibliographystyle{ACM-Reference-Format}
\bibliography{TEG.bib}

\clearpage
\appendix
\section{APPENDIX}

\subsection{Datasets} \label{app:datasets}
In this section, we describe more details on the datasets used for this paper. 
\begin{itemize}[leftmargin=2mm]
    \item \textbf{Cora-full} \cite{corafull} is a citation network where papers are represented as nodes with bag-of-words attributes, and edges represent citation links, while labels indicate the paper topic.
    \item \textbf{Amazon Clothing} \cite{amazon} is a network of products from Amazon's "Clothing, Shoes and Jewelry" category, where nodes are products with descriptions as attributes, connected through "also-viewed" relationships, and labeled by low-level product categories.
    \item \textbf{Amazon Electronics} \cite{amazon} network represents products in the "Electronics" category on Amazon as nodes with descriptions as attributes, connected by "bought-together" relationships, and labeled by low-level product categories.
    \item \textbf{DBLP}  \cite{dblp} network represents papers as nodes connected by citation relationships, with node features derived from paper abstracts and labels assigned based on publication venues.
    \item \textbf{Coauthor-CS} \cite{coauthor} network represents co-authorship relationships among authors, with nodes indicating authors, edges denoting co-authored connections, and labels representing research fields based on paper keywords. 
\end{itemize}

\vspace{-1.3em}

\subsection{Baselines} \label{app:baselines}

Here are descriptions and implementation details of the baseline methods used in the experiments:

\begin{itemize}[leftmargin=2mm]
    \item \textbf{MAML} \cite{maml} optimizes the model parameters to enable rapid adaptation through a few number of fine-tuning steps, utilizing gradients of the support set across meta-tasks. We implemented it with 2 linear layers (hidden dimension: 32) using ReLU activation, a learning rate of 0.001, meta-learning rate of 0.1, and weight decay of 0.0005.
    \item \textbf{ProtoNet} \cite{protonet} learns a set of prototypes for each class and then classifies the query based on Euclidean distance. We implemented it with 2 linear layers (hidden dimension: 128, output dimension: 64) using ReLU activation, a learning rate of 0.001, and weight decay of 0.0005.
    \item \textbf{Meta-GNN} \cite{metagnn} integrates attributed networks with MAML \cite{maml} using GNNs. We implemented it with 5 fine-tuning steps, an update-learning rate of 0.001, meta-learning rate of 0.1, a hidden layer size of 16, dropout rate of 0.5, and weight decay of 0.0005.
    \item \textbf{GPN} \cite{gpn} learns a set of prototypes for each class by using the degree of each node to determine the node's score, which is reflected in the prototypes. We implemented it with 2 graph encoders with a hidden size of 64, a learning rate of 0.001, dropout rate of 0.5, and weight decay of 0.0005.
    \item \textbf{G-Meta} \cite{gmeta} generates node representations by utilizing subgraphs, making it a scalable and inductive meta-learning method for graphs. We set the meta-learning rate to 0.01, update-learning rate to 0.001, 10 update steps, and a 128-dimensional GNN hidden layer.
    \item \textbf{TENT} \cite{tent} generates the graph structure for each class and its GNNs' parameter depending on the class prototypes to address the task variance. We set the learning rate to 0.05, hidden layer size to 16, weight decay to 0.0001, and dropout rate to 0.2.
\end{itemize}

\subsection{Implementation details}
\label{app: Implementation details}

In our experiments, we use a 1-layer GCN implementation ($\mathrm{GCN}_\theta$) with dropout rate of 0.5 to reduce input features to 64 dimensions.
We create 16-dimensional structural features using virtual anchor nodes ($k=16$) for all datasets. 
The model is optimized with Adam using a learning rate of 0.001 and weight decay of 0.0005, while the loss weight coefficient $\gamma$ is set to 0.5. 
The task embedder consists of three main learnable functions as follows:

\begin{itemize}[leftmargin=2mm]
    \item\textbf{Message generating function} ($\phi_m$): Two linear layers with SiLU activation (Inputs $\rightarrow$ Linear ($2d_s+1 \rightarrow 64$) $\rightarrow$ SiLU $\rightarrow$ Linear (64 $\rightarrow$ 64) $\rightarrow$ SiLU $\rightarrow$ Outputs).
    \item\textbf{Message embedding function} ($\phi_l$): Three linear layers with SiLU activation (Inputs $\rightarrow$ Linear (64 $\rightarrow$ 64) $\rightarrow$ SiLU $\rightarrow$ Linear (64 $\rightarrow$ 64) $\rightarrow$ Linear (64 $\rightarrow$ 1) $\rightarrow$ Outputs).
    \item\textbf{Property updating function} ($\phi_s$): Two linear layers with SiLU activation (Inputs $\rightarrow$ Linear ($d_s$+64 $\rightarrow$ 64) $\rightarrow$ SiLU $\rightarrow$ Linear (64 $\rightarrow$ $d_s$) $\rightarrow$ Outputs).
\end{itemize}
In all experiments, our task embedder employs these functions.
We utilize 2-layer task embedders.

\subsection{Complexity Analysis} \label{apx:time}
\vspace{-1ex}
\noindent\subsubsection*{\textbf{Time complexity for Structural feature generation.}} BFS computes single source shortest paths in $O(E+V)$ time, where $E$ is the number of edges and $V$ is the number of nodes. 
Additionally, we only calculate the shortest path distance between specific nodes (i.e., "virtual anchor" and "target"), rather than considering all pairs of nodes.

\noindent\subsubsection*{\textbf{Time complexity for GCN}} We use a GCN for the embedder. The time complexity of a GCN layer is $O(Ed+Vd^2)$, where $d$ represents the latent feature dimension.

\noindent\subsubsection*{\textbf{Time complexity for EGNN + ProtoNet.}} During task-adaptation (EGNN + ProtoNet), a task-specific graph structure is constructed, and pairwise distances between nodes are calculated, resulting in a time complexity of $O(V_{\text{task}}^2d)$, where $V_{\text{task}}$ is the number of nodes in the task. However, we can reduce it to $O(V_{\text{spt}}d)$ by connecting only the support nodes and queries, not all nodes in the task. Thus, the time complexity during task-adaptation is $O(V_{\text{spt}}d)$, which is negligible due to $V_{\text{spt}}<<V$.


In conclusion, the total time complexity of TEG is $O(Ed+Vd^2)$.

\subsection{Effect of Virtual Anchor Nodes} \label{apx:effect_virtual}
\vspace{-1.5ex}
\begin{table}[h]
\centering
\caption{Effect of using virtual anchor node for alleviating no-path-to-reach problem. AC and AE denotes "Amazon Clothing" and "Amazon Electronics", respectively.}
\vspace{-2ex}
\resizebox{\columnwidth}{!}
{
{
\renewcommand{\arraystretch}{1.0}
\begin{tabular}{c||c|cc|cc} \toprule
                   &             & \multicolumn{2}{c|}{\textbf{with virtual anchor nodes}}                                                                                                                                                                         & \multicolumn{2}{c}{\textbf{w.o. virtual anchor nodes}}                                                                                                                                            \\ \cline{2-6}
                   & Dimension   & \# Zero value & Zero ratio & \# Zero value & Zero ratio \\ \hline
Corafull           & 19,793 $\times$ 16 & 544                                                                                                                                            & 0.002                                                           & 15,888                                                                                                                                         & 0.050                                                           \\
AC    & 24,919 $\times$ 16 & 1,280                                                                                                                                          & 0.003                                                           & 66,662                                                                                                                                         & 0.167                                                           \\
AE & 42,318 $\times$ 16 & 9,472                                                                                                                                          & 0.014                                                           & 666,935                                                                                                                                        & 0.985                                                           \\
DBLP               & 40,672 $\times$ 16 & 0                                                                                                                                              & 0.000                                                           & 352                                                                                                                                            & 0.001   \\ \bottomrule                                                       
\end{tabular}
}
}
\label{apx:virtual}
\end{table}
\vspace{-1em}
When calculating the shortest path distance $d^{sp}(v,u)$ between two different nodes $v$ and $u$, if node $v$ is unable to reach the target node $u$, the value of $d^{sp}(v,u)$ becomes infinity, resulting in $s(u,v)$ being equal to 0.
Having an excessive number of 0s in the feature can result in a lack of sufficient structural information. 
To address this issue, we mitigate the problem by introducing virtual anchor nodes.
Specifically, in Table~\ref{apx:virtual}, we observe that when structural features are generated based on randomly selected anchor nodes (i.e., w.o. virtual anchor nodes), particularly in the case of Amazon Electronics, a severe problem arises where 98.5\% of the target nodes do not have any path to reach the anchor node. 
In other words, alternative calculation methods that utilize the original graph, such as Random Walks and PGNN \cite{pgnn}, would also suffer from the same problem, as they do not consider virtual anchor nodes. 
However, our approach (i.e., with virtual anchor nodes) effectively mitigates this problem, reducing the zero ratio from 98.5\% to 1.4\% in Amazon Electronics.
\vspace{-1mm}

\subsection{Additional Experiments} \label{apx:experiments}
\begin{figure}[H] 
\vspace{-1.em}
\begin{center}
\includegraphics[width=1.0\linewidth,height=2.5cm]{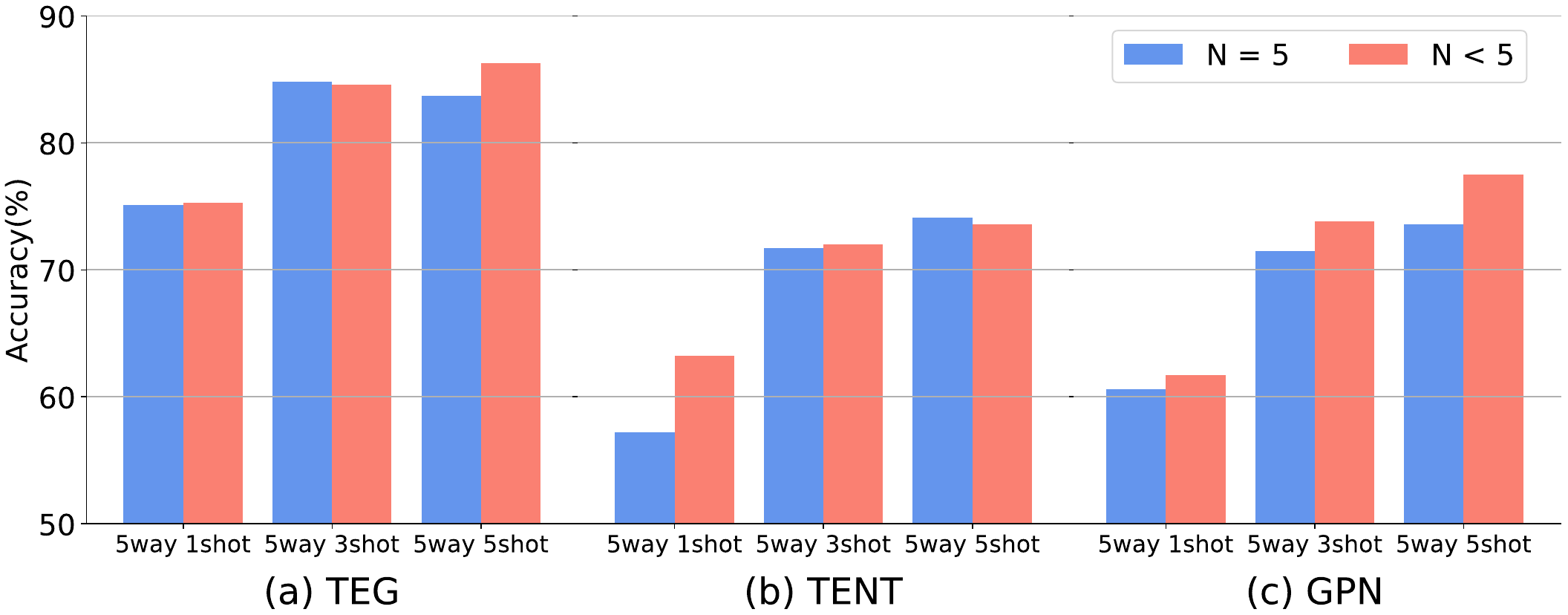}
\end{center}
\vspace{-2mm}

\caption{Effect of reducing \textit{way} in training on 5-way $K$-shot evaluation setting. Bar denotes mean accuracy of 5-way K-shot meta-testing tasks, predicted by model trained on N-way K-shot meta-training tasks using Coauthor-CS dataset.}
\label{fig:diversity1}
\vspace{-3mm}
\end{figure}
\subsubsection*{\textbf{Episode Composition for Low Diversity Training Dataset}}

To address low meta-task diversity, we propose effective training techniques for metric-based methods. 
The learned metric in these methods is unaffected by the number of classes (\textit{way}), allowing flexibility in adapting to changes during training and testing. 
Existing few-shot learning methods \cite{protonet} suggest that training with higher \textit{way} settings can improve performance, but this requires a large number of labeled classes, which can be costly in the graph domain. 
Instead, we argue that training with lower \textit{way} settings can be more appropriate and beneficial as it generates diverse class combinations, enhancing the diversity of training meta-tasks. 
Our experiments on the Coauthor-CS dataset demonstrate that training with lower \textit{way} settings can achieve comparable performance to higher \textit{way} settings. 
In Figure \ref{fig:diversity1}, we compare the results of \textit{training with $N$-way ($N=5$) settings} and \textit{training with $N$-way ($N<5$) settings}.
The performance of the $N<5$ settings denotes the average performance being achieved between 2/3/4-way settings.
While our model shows strong generalization capacity in low-diversity meta-tasks, TENT and GPN exhibit a steeper performance gain. 
Overall, our findings highlight the effectiveness of lower \textit{way} settings for training in scenarios with limited diversity and the generalization capabilities of our model in low-diversity meta-task scenarios.

\subsection{Equivariance Proof} \label{apx:equi-proof}

We prove that our task embedder, based on EGNN \cite{egnn}, is translation-equivariant on a semantic feature, $\mathrm{\textbf{h}}^{(l)}_i$ (i.e., coordinates of a node $i$), for any translation vector $g\in\mathbb{R}^{d_l}$, and rotation/reflection-equivariant on $\mathrm{\textbf{h}}^{(l)}_i$ for any orthogonal matrix $\textbf{Q}\in\mathbb{R}^{d_l\times d_l}$.
To this end, we will formally prove that our task embedder satisfies:
\begin{equation} \label{eq:proof}
    \mathbf{\textbf{h}}^{(l),\lambda+1}_i\textbf{Q}+g, \mathrm{\textbf{h}}^{(s),\lambda+1}_i=\phi(\mathbf{\textbf{h}}^{(l),\lambda}_i\textbf{Q}+g, \mathrm{\textbf{h}}^{(s),\lambda}_i).
\end{equation}

Table \ref{tbl:equiproof} summarizes the task embedder process and indicates each step's equivariance/invariance to transformations (i.e., translation, rotation, reflection).
\vspace{-1em}
\begin{table}[h]
\caption{Process of \textit{task-adaptation} in a task embedder}
\vspace{-2.5mm}
\label{tbl:equiproof}
\resizebox{\columnwidth}{!}
{
{
\renewcommand{\arraystretch}{1.2}
\begin{tabular}{c||c|c|c||c} \toprule
Step & Inputs & Function & Outputs & Equivariance \\ \hline\hline
1    & $\mathrm{\mathbf{h}}^{(s),\lambda}_i, \mathrm{\mathbf{h}}^{(s),\lambda}_j, \mathrm{\mathbf{h}}^{(l),\lambda}_i, \mathrm{\mathbf{h}}^{(l),\lambda}_j$      & $ \phi_m(\mathrm{\mathbf{h}}^{(s),\lambda}_i, \mathrm{\mathbf{h}}^{(s),\lambda}_j, \lVert \mathrm{\mathbf{h}}^{(l),\lambda}_i-\mathrm{\mathbf{h}}^{(l),\lambda}_j \rVert ^2)$        & $\mathrm{\mathbf{m}}_{ij}$       & Invariance   \\ \hline
2    & $\mathrm{\mathbf{h}}^{(l),\lambda}_i, \mathrm{\mathbf{h}}^{(l),\lambda}_j, m_{ij}$      & $\mathrm{\mathbf{h}}^{(l),\lambda}_i+\frac{1}{C}\sum_{j\neq i}(\mathrm{\mathbf{h}}^{(l),\lambda}_i-\mathrm{\mathbf{h}}^{(l),\lambda}_j)\phi_l(\mathrm{\mathbf{m}}_{ij})$        & $\mathrm{\mathbf{h}}^{(l),\lambda+1}_i$       & Equivariance \\ \hline
3    & $\mathrm{\mathbf{m}}_{ij}$      & $\sum_{j \in \mathcal{N}(i)}\mathrm{\mathbf{m}}_{ij}$        & $\mathrm{\mathbf{m}}_{i}$       & Invariance   \\ \hline
4    & $\mathrm{\mathbf{h}}^{(s),\lambda}_i, m_i$      & $\phi_s(\mathrm{\mathbf{h}}^{(s),\lambda}_i, 
\mathrm{\mathbf{m}}_{i})$        & $\mathrm{\mathbf{h}}^{(s),\lambda+1}_i$       & Invariance  \\ \bottomrule
\end{tabular}
}
}
\end{table}
\vspace{-1em}

\noindent
\textbf{Step 1.}
We first generate the message between the support set and queries in Step 1.
Structural feature, $\mathrm{\textbf{h}}^{(s)}_i$ (i.e., property of a node $i$), remain invariant under any transformation.
Relative squared distance between two instances also stays invariant as it is invariant to translations, rotations, and reflections.

\noindent 
\textit{\textbf{Deviation.}}
\vspace{-3mm}

\begin{equation*}
\begin{aligned}
    \lVert [\mathrm{\mathbf{h}}^{(l),\lambda}_i+g]-[\mathrm{\mathbf{h}}^{(l),\lambda}_j+g] \rVert ^2=\lVert \mathrm{\mathbf{h}}^{(l),\lambda}_i-\mathrm{\mathbf{h}}^{(l),\lambda}_j \rVert ^2 \\ 
    \cdots \text{translation invariant}&\\
    \lVert \mathrm{\mathbf{h}}^{(l),\lambda}_i\textbf{Q}-\mathrm{\mathbf{h}}^{(l),\lambda}_j\textbf{Q} \rVert ^2=(\mathrm{\mathbf{h}}^{(l),\lambda}_i-\mathrm{\mathbf{h}}^{(l),\lambda}_j)^\mathrm{T}\textbf{Q}^\mathrm{T}\textbf{Q}(\mathrm{\mathbf{h}}^{(l),\lambda}_i-\mathrm{\mathbf{h}}^{(l),\lambda}_j)&\\= (\mathrm{\mathbf{h}}^{(l),\lambda}_i-\mathrm{\mathbf{h}}^{(l),\lambda}_j)^\mathrm{T}\textbf{I}(\mathrm{\mathbf{h}}^{(l),\lambda}_i-\mathrm{\mathbf{h}}^{(l),\lambda}_j)& \\=\lVert\mathrm{\mathbf{h}}^{(l),\lambda}_i-\mathrm{\mathbf{h}}^{(l),\lambda}_j \rVert ^2& \\\cdots \text{rotation, reflection invariant}&\\
\textstyle\therefore\mathrm{\mathbf{m}}_{ij}=\phi_m(\mathrm{\mathbf{h}}^{(s),\lambda}_i, \mathrm{\mathbf{h}}^{(s),\lambda}_j, \lVert [\mathrm{\mathbf{h}}^{(l),\lambda}_i\textbf{Q}+g]-[\mathrm{\mathbf{h}}^{(l),\lambda}_j\textbf{Q}+g] \rVert ^2)&\\\textstyle=\phi_m(\mathrm{\mathbf{h}}^{(s),\lambda}_i, \mathrm{\mathbf{h}}^{(s),\lambda}_j, \lVert \mathrm{\mathbf{h}}^{(l),\lambda}_i-\mathrm{\mathbf{h}}^{(l),\lambda}_j \rVert ^2)&\\\cdots \text{translation, rotation, reflection invariant}&
\end{aligned}
\end{equation*}
\vspace{-3mm}

\noindent
\textbf{Step 2.}
Step 2 updates the coordinates based on the message and differences between the coordinates. 
As we have shown in Step 1, $\mathrm{\mathbf{m}}_{ij}$ is invariant to transformations, and thus, $\phi_l(\mathrm{\mathbf{m}}_{ij})$ also maintains its invariance to transformations.
Since the differences between the coordinates preserve the transformations, the updated coordinates remain equivariant, meaning input transformations result in equivalent transformations in the output.

\noindent
\textit{\textbf{Deviation.}}
\vspace{-3mm}

\begin{equation*}
\begin{aligned}
    \textstyle [\mathrm{\mathbf{h}}^{(l),\lambda}_i\textbf{Q}+g]+\frac{1}{C}\sum_{j\neq i}([\mathrm{\mathbf{h}}^{(l),\lambda}_i\textbf{Q}+g]-[\mathrm{\mathbf{h}}^{(l),\lambda}_j\textbf{Q}+g])\phi_l(\mathrm{\mathbf{m}}_{ij})\\=\textstyle\mathrm{\mathbf{h}}^{(l),\lambda}_i\textbf{Q}+g+\frac{1}{C}\sum_{j\neq i}((\mathrm{\mathbf{h}}^{(l),\lambda}_i+g-\mathrm{\mathbf{h}}^{(l),\lambda}_j-g)\textbf{Q})\phi_l(\mathrm{\mathbf{m}}_{ij})\\\textstyle=(\mathrm{\mathbf{h}}^{(l),\lambda}_i+\frac{1}{C}\sum_{j\neq i}(\mathrm{\mathbf{h}}^{(l),\lambda}_i-\mathrm{\mathbf{h}}^{(l),\lambda}_j)\phi_l(\mathrm{\mathbf{m}}_{ij}))\textbf{Q}+g\\\textstyle=\mathrm{\mathbf{h}}^{(l),\lambda+1}_j\textbf{Q}+g\\
    \cdots \text{translation, rotation, reflection equivariant}
\end{aligned}
\end{equation*}

\vspace{-1em}
\begin{equation*}
\begin{aligned}
    \textstyle\therefore \mathrm{\mathbf{h}}^{(l),\lambda+1}_j&\textbf{Q}+g=\\&\textstyle[\mathrm{\mathbf{h}}^{(l),\lambda}_i\textbf{Q}+g]+\frac{1}{C}\sum_{j\neq i}([\mathrm{\mathbf{h}}^{(l),\lambda}_i\textbf{Q}+g]-[\mathrm{\mathbf{h}}^{(l),\lambda}_j\textbf{Q}+g])\phi_l(\mathrm{\mathbf{m}}_{ij})\\
     & \quad\quad\quad\quad \cdots \text{translation, rotation, reflection equivariant}
\end{aligned}
\end{equation*}

\noindent
\textbf{Step 3, 4.} 
The inputs in Step 3 and 4 are always invariant to transformations because the message $\mathrm{\mathbf{m}}_{ij}$ was shown to be invariant in Step 2, and the meta-task's structural features are inherently invariant to transformations.
Therefore, it is obvious that the outputs of the functions in Step 3 and 4 are invariant to translation, rotation, and reflection. Hence, our task embedder satisfies Equation~\ref{eq:proof}.


\end{document}